%% file: neurips_2024.tex
\documentclass{article}

\usepackage[preprint,nonatbib]{neurips_2024}

\usepackage[utf8]{inputenc} 
\usepackage[T1]{fontenc}    
\usepackage{hyperref}       
\usepackage{url}            
\usepackage{booktabs}       
\usepackage{amsfonts}       
\usepackage{nicefrac}       
\usepackage{microtype}      
\usepackage{xcolor}         
\usepackage{todonotes}
\usepackage{amsmath}
\usepackage{amssymb}
\usepackage{amsthm}
\usepackage{pifont}
\usepackage{wrapfig}

\usepackage{longtable}
\usepackage{array}

\allowdisplaybreaks

\definecolor{darkgreen}{rgb}{0.0, 0.6, 0.0}

\title{OccamLLM: Fast and Exact Language Model Arithmetic in a Single Step}

\author{%
  Owen Dugan\thanks{Equal contribution}\;\;\thanks{Corresponding author}\\
  Department of Physics\\
  Massachusetts Institute of Technology\\
  Cambridge, MA \\
  \texttt{odugan@mit.edu} \\
  \And
  Donato M. Jiménez-Benetó\footnotemark[1]\\
  Department of Physics\\
  Massachusetts Institute of Technology\\
  Cambridge, MA \\
  \texttt{donatojb@mit.edu} \\
  \And
  Charlotte Loh \\
  Department of EECS\\
  Massachusetts Institute of Technology\\
  Cambridge, MA \\
  \texttt{cloh@mit.edu} \\
  \And
  Zhuo Chen\\
  Department of Physics\\
  Massachusetts Institute of Technology\\
  Cambridge, MA \\
  \texttt{chenzhuo@mit.edu} \\
  \And
  Rumen Dangovski\\
  Department of EECS\\
  Massachusetts Institute of Technology\\
  Cambridge, MA \\
  \texttt{rumenrd@mit.edu} \\
  \And
  Marin Solja\v{c}i\'{c}\\
  Department of Physics\\
  Massachusetts Institute of Technology\\
  Cambridge, MA \\
  \texttt{soljacic@mit.edu} \\
}

\begin{document}

\maketitle

\begin{abstract}
    Despite significant advancements in text generation and reasoning, Large Language Models (LLMs) still face challenges in accurately performing complex arithmetic operations. Language model systems often enable LLMs to generate code for arithmetic operations to achieve accurate calculations. However, this approach compromises speed and security, and fine-tuning risks the language model losing prior capabilities. We propose a framework that enables exact arithmetic in \textit{a single autoregressive step}, providing faster, more secure, and more interpretable LLM systems with arithmetic capabilities. We use the hidden states of a LLM to control a symbolic architecture that performs arithmetic. Our implementation using Llama 3 with OccamNet as a symbolic model (OccamLlama) achieves 100\% accuracy on single arithmetic operations ($+,-,\times,\div,\sin{},\cos{},\log{},\exp{},\sqrt{}$), outperforming GPT 4o with and without a code interpreter. Furthermore, OccamLlama outperforms GPT 4o with and without a code interpreter on average across a range of mathematical problem solving benchmarks, demonstrating that OccamLLMs can excel in arithmetic tasks, even surpassing much larger models. We will make our code public shortly.
\end{abstract}

\section{Introduction}
Since the release of GPT 3, Large Language Models (LLMs) have dramatically improved in their text generation and reasoning capabilities. This has enabled success in downstream applications including machine translation \cite{translation_1, translation_multi}, sentiment analysis \cite{sa_1, sa_2, sa_3}, and interactive dialogue generation \cite{gpt4}, with language models even surpassing human experts on some academic benchmarks that require reading comprehension, reasoning and coding \cite{gemini}.
However even industry-leading LLMs such as GPT 4 cannot reach 100\% accuracy on simple arithmetic \cite{math401}, limiting their ability to perform basic mathematical tasks. This hinders potential applications of LLMs ranging from chat-bot physics tutors to LLM-powered automated research that could accelerate scientific discovery and technological innovation. The poor arithmetic performance of LLMs is particularly acute for small LLM agents, limiting their usage in smartphone or in multi-agent applications.

To enable accurate calculations, language model systems often resort to running code written by a LLM. However, this comes at the cost of speed; the model must perform multiple autoregressive steps to generate code that performs the appropriate arithmetic operations. This increased decoding time may negatively impact applications such as multi-agent workflows \cite{multiagent_1, multiagent_2} where speed is essential. At the same time, code-based LLM arithmetic mechanisms may increase system vulnerability by providing a mechanism for arbitrary LLM-generated code execution.

We propose an alternative, a framework which enables exact and interpretable LLM arithmetic in \textit{a single autoregressive step}, providing faster and more secure arithmetic capabilities in LLM systems. Our framework uses the hidden states of a LLM to control a symbolic architecture that performs arithmetic. Although our method can in principle work with any symbolic architecture, in this paper we use an interpretable neurosymbolic architecture known as OccamNet \cite{OccamNet, OccamNetSocialSci} because of its interpretability and scalability. Therefore, we term our method OccamLLM, or OccamLlama when using a Llama model as the LLM.

Our core contributions are as follows:
\begin{enumerate}
    \item We develop a framework for exact and interpretable LLM arithmetic in a single autoregressive step without catastrophic forgetting \cite{catastrophic} or vulnerability from code generation. We explore how to train OccamLlama, including data generation, decoder architecture, and loss function.
    \item We benchmark OccamLlama on arithmetic tasks, demonstrating that OccamLlama achieves 100\% accuracy on arbitrary single arithmetic operations ($+,-,\times,\div,\sin{},\cos{},\log{},\exp{},\sqrt{}$), more than double the accuracy of GPT 4o. OccamLlama performs slightly better than GPT 4o with Code Interpreter while answering in on average more than 50x fewer generation tokens.
    \item We benchmark on mathematical problem solving tasks, showing that OccamLlama can sustain long generations. OccamLlama outperforms both GPT 4o and GPT 4o with code interpreter on average across the benchmarks we tested.
\end{enumerate}

\begin{figure}[t]
    \centering
    \includegraphics[width=\linewidth]{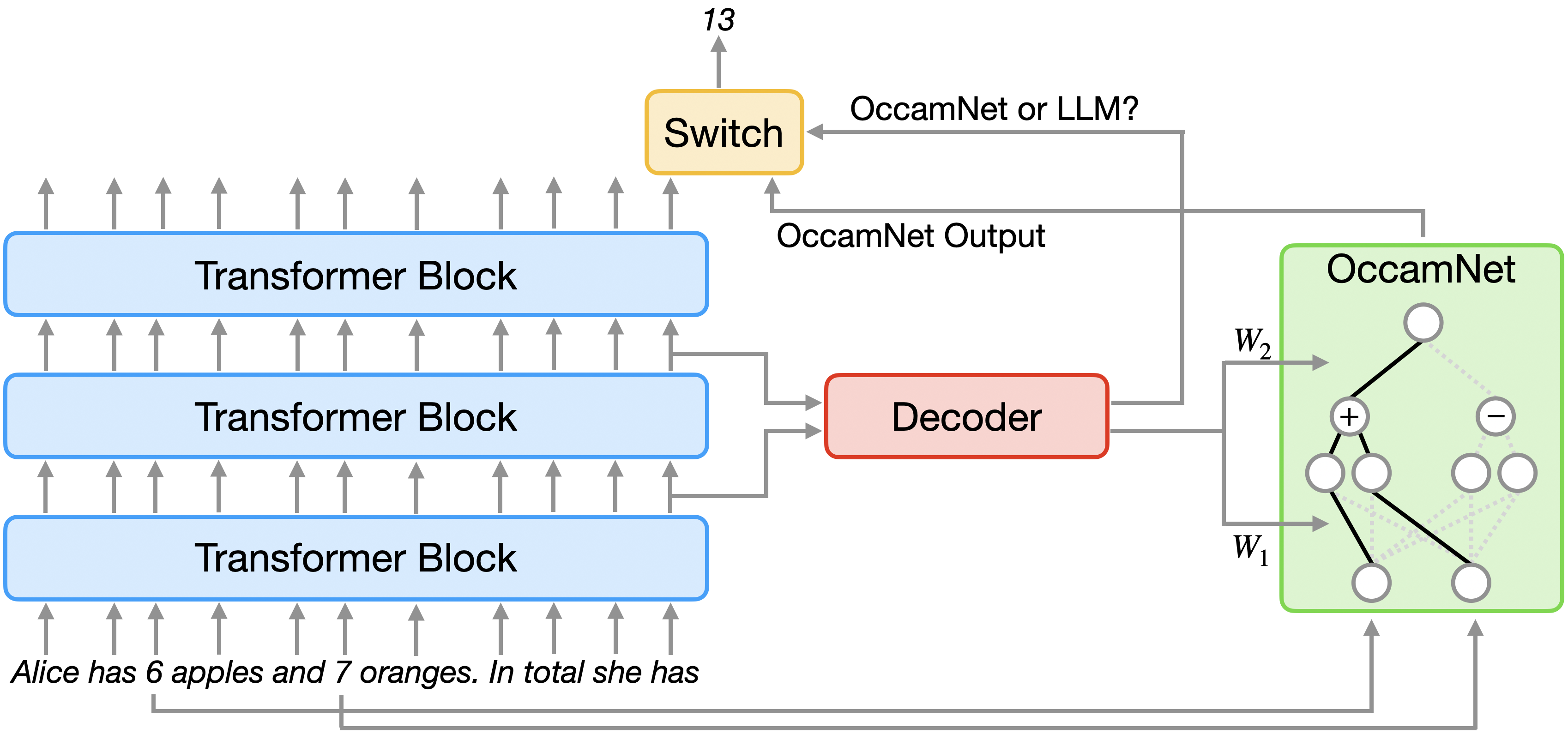}
    \caption{The OccamLLM system. For each autoregressive step, the language model hidden states for that token are fed into a decoder block which assigns weights to OccamNet. The system feeds the most recent numbers from the text into OccamNet, which then evaluates the sparse function specified by its weights. The decoder then determines whether to use the LLM output or the OccamNet output.}
    \label{fig: occamllama}
\end{figure} 

\section{Related Work}

\begin{table}[t]
\centering
\caption{OccamLLM is the only approach to improving the arithmetic capabilities of a pretrained LLM which 1) enables single-pass arithmetic, 2) does not risk catastrophic forgetting from finetuning, 3) does not require arbitrary code execution, and 4) provides an interpretable process.}
\begin{tabular}{lcccc}
\toprule
&  & No Catastrophic & No Arbitrary & \\ 
& Single Pass & Forgetting & Code Execution & Interpretable\\ 
\cmidrule(r){2-2}
\cmidrule(r){3-3}
\cmidrule(r){4-4}
\cmidrule(r){5-5}
Fine Tuning & \textcolor{darkgreen}{\ding{51}} & \textcolor{red}{\ding{55}} & \textcolor{darkgreen}{\ding{51}} & \textcolor{red}{\ding{55}} \\ 
Tool Use & \textcolor{red}{\ding{55}} & \textcolor{red}{\ding{55}} & \textcolor{red}{\ding{55}} & \textcolor{darkgreen}{\ding{51}} \\ 
\textbf{OccamLLM} & \textcolor{darkgreen}{\ding{51}} & \textcolor{darkgreen}{\ding{51}} & \textcolor{darkgreen}{\ding{51}} & \textcolor{darkgreen}{\ding{51}} \\ 
\bottomrule
\end{tabular}
\vspace{-1em}
\label{tab:comparison}
\end{table}

\paragraph{Arithmetic Performance in LLMs.}

Prior research has trained models on synthetic data, finding that such models can achieve near-perfect accuracy on addition \cite{2023arXiv230801154M, 2023arXiv230903241Y}, subtraction \cite{2023arXiv230903241Y}, multiplication \cite{2023arXiv230801154M, 2023arXiv230903241Y}, division \cite{2023arXiv230903241Y}, and raising to powers \cite{2023arXiv230903241Y}. These prior models have been tested only on arithmetic datasets, so their generality has not been assessed. Other work focuses on finetuning LLMs which are already trained on large amounts of general-purpose data on math datasets. Both full-parameter \cite{ft_full, wang2023mathcoder} and parameter-efficient (PEFT) \cite{ft_adapters} finetuning strategies have been applied. However, finetuning on a single dataset carries the risk of catastrophic forgetting of an LLM's previously acquired linguistic skills \cite{french1999catastrophic}. While PEFT techniques have been shown to partially mitigate this effect, this area is still one of active research \cite{peft_1, peft_2}.

\paragraph{LLMs with Tool Use.}
Another thrust of prior research has focused on LLM tool use, which we believe is most directly related to our methods. \textit{Calc-X}~\cite{kadlčík2023calcx} introduces a technique to offload arithmetic computations to an external tool like a calculator. 
The authors curated a large dataset of arithmetic problems and trained a language model that learns to interact with a calculator through the use of tags to signify the calling of the external tool. 
Several other works~\cite{komeili-etal-2022-internet,nakano2022webgpt,thoppilan2022lamda} follow a similar idea, using crowd workers to annotate tool calls and using this data to train language models to interact with external tools such as a web searching tool, a calculator, or a translation system. 
These approaches can be prohibitively expensive in annotation costs; \textit{Toolformer}~\cite{schick2023toolformer} overcomes this cost by using in-context learning and a language model to generate datasets containing the necessary `API' tool calls via a self-supervised loss.
Further, the above methods all require finetuning of the LLM, placing the LLM at risk of losing generality and its original language modelling abilities through catastrophic forgetting. 
In contrast, our approach does not involve training the language model. Our `external tool' is a symbolic model which can be trained to correctly use the hidden states of the language model to perform the required arithmetic computations. The language model is kept frozen throughout this process.
Unlike other tool-calling approaches, where the cost of data annotation to train for tool-calling interaction can be prohibitively expensive, in our method, each task only requires manually annotating tens of prompts, a high annotation efficiency.
Other prior methods leverage prompt engineering to improve arithmetic performance of LLMs; this is done either through chain-of-thought~\cite{wei2023chainofthought}, or to encourage LLMs to use a code interpreter~\cite{gao2023pal, chen2023program, zhou2023solving}. 
Contrary to these methods, our approach does not use code kernels; this provides several advantages: 1) it enables tool use without expending compute on autoregressive steps for token generation, and 2) it avoids running potentially incorrect or malicious code generated by language models.

\section{Methods}

\subsection{OccamLLM: Combining a Language Model with a Symbolic Model}

In short, the OccamLLM system combines a language model with a symbolic model, namely OccamNet, that can perform arithmetic operations like addition and subtraction. For each token, the corresponding internal hidden states of the language model are fed into a decoder module which initializes the symbolic model such that it executes the operation required by the task described in the input text. A string parser feeds the necessary numbers from the text into OccamNet, which evaluates the desired expression. Finally, a decoder determines whether to use the language model output or the OccamNet output for generating the next token. 

In the example shown in Figure \ref{fig: occamllama}, a decoder determines how to initialize OccamNet from the language model hidden states, choosing to have OccamNet perform addition. The text parser then feeds the numbers $6$ and $7$ into OccamNet, which adds the numbers, returning $13.$ Finally, a decoder decides to use the OccamNet output instead of the language model output, so the system outputs 13. The new sentence, including the 13, is tokenized and fed back to the LLM to continue autoregressive generation. The language model might later generate ``Since she ate two apples, she now has,'' at which point the switch will again trigger OccamNet, this time implementing $13-2$ and returning $11.$

In the subsections below, we describe the OccamLLM system which from our experiments we find to be most performant, even oupterforming GPT 4o in several benchmarks. For an analysis of alternate architectures and losses, see Appendix \ref{app: alternative architectures}.
\looseness=-1

\subsubsection{OccamNet}
OccamNet is a symbolic architecture that provides an interpretable way of parametrizing probability distributions over a space of functions \cite{OccamNet}. We leave a more thorough explanation of OccamNet to \cite{OccamNet} and Appendix \ref{app: background on occamnet}, describing only the relevant components here.

\begin{figure}
    \centering
    \includegraphics[width=\linewidth]{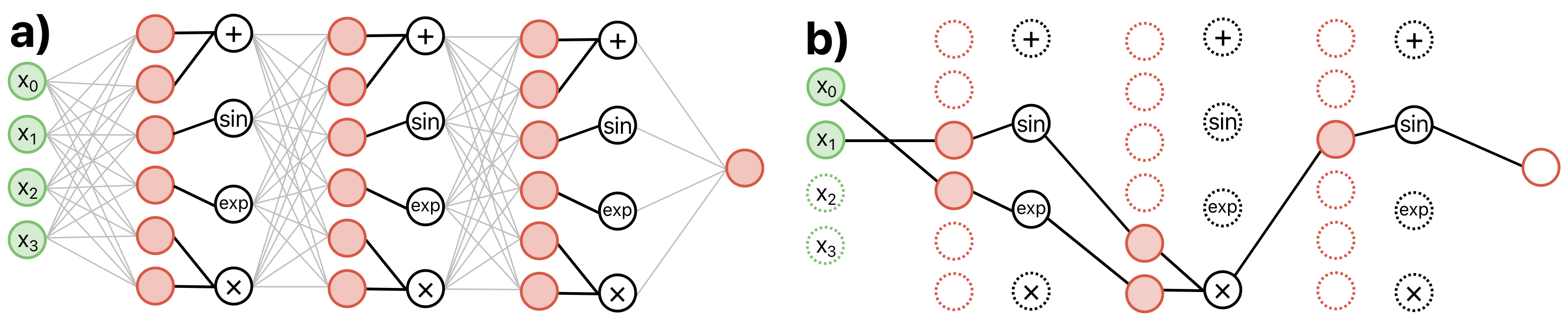}
    \caption{a) A schematic of the OccamNet architecture, with softmax layers in grey and their outputs in red. b) A Directed Acyclic Graph (DAG) (with edges not connected to the output removed for clarity) formed by sampling from OccamNet. This DAG corresponds to the function $\sin(\sin(x_1)\cdot \exp(x_0)).$ Modified from \cite{OccamNet}.}
    \label{fig: occamnet}
\end{figure}

An $l$-layer OccamNet with primitives $\mathcal{P}$ and $n$ inputs is an architecture that defines a probability distribution over the space of functions representable as compositions of the primitives in $\mathcal{P}$ up to depth $l.$ For example, a two-layer OccamNet with primitives $\mathcal{P} = \{\sin, \cos\}$ and one input represents a probability distribution over the set
\begin{equation*}
    \mathcal{F} = \{x, \sin(x), \cos(x), \sin(\sin(x)), \sin(\cos(x)), \cos(\sin(x)), \sin(\sin(x))\}.
\end{equation*}
OccamNet has the structure of an $n$-input, $l$-internal-activation-layer multilayer perceptron with the biases removed and the activations in each layer replaced by the primitives $\mathcal{P},$ as shown in Figure \ref{fig: occamnet}a. Activation functions may have multiple inputs. We rename the linear layers \textit{softmax layers}, denote the weights of the $i$th softmax layer as $\mathbf{W}^{(i)}$, and denote the combined weights of OccamNet as $\mathbf{W}.$

We define the probability distribution which OccamNet parametrizes by specifying how to sample from it. For each softmax layer output node (shown in red in Figure \ref{fig: occamnet}), we select a single connection to that node from a softmax layer input node by sampling from the distribution given by the softmax of the weights of the connections to the different inputs. This process produces a directed acyclic graph (DAG) defining a computational path through the OccamNet activations, such as the one shown in Figure \ref{fig: occamnet}b. In this way, each DAG represents a function on the inputs of OccamNet.

To ensure that OccamNet can represent all possible compositions of functions in $\mathcal{P}$ up to depth $l,$ we include the following modifications to the OccamNet architecture: 1) for each softmax layer, we concatenate its inputs with the previous softmax layer's inputs to enable the representation of functions with fewer than $l$ compositions, and 2) we repeat primitives in the $i$th activation layer $A^{l-i}$ times, where $A$ is the maximum number of inputs of any of the primitives, to ensure that a sufficient number of each primitive is available at each layer. We refer to this modified architecture as \textit{complete OccamNet} as it can represent the complete set of desired functions. The resulting architecture is shown in Figure \ref{fig:mesh2} in the appendix.

In principle, OccamLLM can work with any symbolic model, i.e., any model that can parameterize a set of symbolic functions or a distribution over such functions. We choose OccamNet as opposed to, for example, a transformer \cite{attentionAllYouNeed} or recurrent neural network \cite{rnns}, for two reasons: 1) OccamNet is interpretable, which we hypothesize makes controlling OccamNet an easier task for a decoder to learn, and 2) OccamNet is parallelizable over multiple samples, allowing for scalable training.

\subsubsection{OccamLLM Decoder} \label{subsec: occamllm decoder}

The OccamLLM decoder takes the hidden states of a language model and outputs an initialization for OccamNet. This gives the LLM control over which function to apply on the inputs. The decoder acts on each input token separately, producing a different OccamNet initialization for each. Therefore, the arithmetic operations predicted may change along an input sequence, allowing OccamNet's use for different computations in a single multi-token generation. This is crucial in multi-step reasoning scenarios where OccamNet is employed several times for different purposes.

Many decoder architectures are possible. We choose to parameterize the weights of each softmax layer of OccamNet independently, as $(\mathbf{W}^{(1)},\ldots,\mathbf{W}^{(l)}) = (\text{Decoder}_1(\mathbf{h}),\ldots,\text{Decoder}_l(\mathbf{h})),$ where $\mathbf{h}$ are the hidden states of the language model. We choose
\begin{equation}
    \text{Decoder}_i(\mathbf{h}) = \text{MLP}_i\left(\sum_{j=1}^L w_{i,j}\mathbf{h}_j\right) + \mathbf{W}^{*(i)}
\end{equation}
where $\mathbf{h}_j$ are the hidden states of the $j$th layer of the language model, $w_{i,j}$ are trainable weights, $\text{MLP}_i$ are two-layer multilayer perceptrons (MLPs), and $\mathbf{W}^{*(i)}$ are untrained weights which initialize all functions to have approximately equal probabilities according to the initialization scheme described in \cite{OccamNet} and explained in Appendix \ref{sec:initialization}.

\subsubsection{OccamLLM Switch} \label{subsec: occamllm switch}
We similarly train a decoder for a switch that, for each input token, is fed the hidden states of the language model and selects whether to use the output of OccamNet or the output of the language model. The decoder outputs a single number from 0 to 1, where all numbers less than or equal to 0.5 correspond to using the output of the language model and all numbers greater than 0.5 correspond to using the output of OccamNet. We choose the following architecture for the switch decoder:
\begin{equation}
    \text{Decoder}_{\text{switch}}(\mathbf{h}) = \text{sigmoid}\left(\text{MLP}_{\text{switch}}\left(\sum_{j=1}^L w_{\text{switch},j}\mathbf{h}_j\right)\right).
\end{equation}

\subsection{Data Generation}\label{sec: Data Generation}

We create synthetic datasets to train the OccamLLM decoders, which contain instruction prompts for diverse arithmetic tasks. To generate datasets of arbitrary size, we create prompts with placeholders for numbers. Each prompt includes a question with number placeholders, the sampling value range for each number, and a function that computes the answer to the query given the sampled input numbers. The prompts fall into two main categories: purely arithmetic tasks and reasoning problems.

Purely arithmetic prompts are formed by expressions including only symbols, without any natural language added, such as ``$3 + 85 = $.'' We create prompts using the following operations: $+(\cdot, \cdot)$, $-(\cdot, \cdot)$, $\times(\cdot, \cdot)$, $\div(\cdot, \cdot)$, $\text{sqrt}(\cdot)$, $\text{power}(\cdot, \cdot)$, $\log_e(\cdot)$, $\exp(\cdot)$, $\sin(\cdot)$, and $\cos(\cdot)$.

We also include word problems that require one or two reasoning steps. We generated 150 single step word problems and 40 multi-step reasoning problems which we modified from examples in the MultiArith training dataset \cite{multiarith}.

\subsubsection{OccamNet Decoder Training Data} \label{subsec: decoder data}

For training the decoder that controls the weights of OccamNet, we created two types of examples, single queries and concatenated queries. For single queries, we select a single prompt from the problems generated as discussed in Section \ref{sec: Data Generation}. We use the Llama 3 Instruct chat template and fill in the query as the user input and the result as the assistant response, prepending ``Answer = '' to the later in randomly selected samples (see Appendix  \ref{app: dataset occamnet} for further details). For the concatenated queries of examples, we select a random number of prompts and concatenate the query-response pairs without using the Llama 3 Instruct chat template. The OccamNet decoder is trained to predict only the results of the last query in the sequence. This strategy helps OccamLLM to learn which operation to perform without becoming confused by earlier text, which is useful for continuous generation. To create the training dataset, each example is sampled by first randomly selecting whether to create a single or concatenated query, then randomly selecting the type(s) of prompt(s) used, and finally randomly sampling the input values from the range corresponding to each selected prompt.

\subsubsection{OccamLLM Switch Training Data}  \label{subsec: switch data}

To train the switch, we generate examples of possible LLM outputs for given input expressions and label the outputs with sequences of 0s or 1s corresponding to whether the language model or the OccamNet output should be used for the next token. Some examples correspond to the prompts described in Section \ref{sec: Data Generation}. For such examples, the LLM output is set to ``The answer is'' or ``Answer = '' and the label sequence is all 0s with a 1 at the last token to indicate the system should use OccamNet only to compute the answer. We also manually created and labeled several other examples for diverse scenarios to explicitly teach the system in which cases it should or should not use OccamNet (see Appendix \ref{app: dataset switch} for further details).

To create the training dataset, we concatenate a random number of the above user input - assistant output pairs in a conversational fashion, using the Llama 3 Instruct chat template.

\subsection{OccamLLM Training}

We train the OccamLLM decoder and the switch separately, as they do not share weights. In all cases, the weights of the LLM are kept frozen. In the first step, we train the system to predict the answer to examples generated by the method explained in Section \ref{subsec: decoder data}. The OccamNet decoder processes the hidden states corresponding to the last token of the response and sets the weights of OccamNet such that the correct arithmetic expression is sampled. In this step, we use a rescaled REINFORCE \cite{reinforce} loss, which can also be interpreted as a Monte-Carlo estimate of the cross-entropy loss (see Appendix \ref{app: alt losses}):
\begin{equation}\label{eq: Loss}
    \mathcal{L}(x,y;W) = - \frac{\sum_{f\sim p_W} R(f(x),y) \log p_W[f]}{\sum_{f\sim p_W} R(f(x),y)},
\end{equation}
where $p_W[f] \equiv \text{ON}(f; \text{Decoder}_W(\mathbf{h}(x)))$ is the probability distribution represented by the decoder-initialized OccamNet.

Minimizing this loss steers the decoder towards assigning higher probabilities to the functions that maximize the reward $R(f(x),y)$, which measures the similarity between the correct answer $y$ and the prediction of OccamNet $f(x)$. We find setting $R(f(x),y)=1$ if $f(x) = y$, and 0 otherwise, most effective. We discuss the OccamNet loss in more detail in Appendix \ref{app: alternative architectures}.

The second step involves training the decoder to route the outputs to OccamNet when needed. We train the switch decoder alone, freezing the weights of the OccamNet decoder of the previous step and minimizing the binary cross-entropy loss between the switch output and the desired output for each token. The OccamLLM switch decoder learns when to route the output to OccamNet in diverse contexts.
\looseness-1

\section{Experiments}\label{sec: experiments}

For all OccamLLM results, we use Llama 3 8B Instruct and Llama 3 70B Instruct \cite{llama3modelcard} as the underlying language models. As such, we call our models OccamLlama 8B and OccamLlama 70B, respectively. We use a 1 layer Complete OccamNet with primitives \begin{align*}
    \mathcal{P} = \{+(\cdot, \cdot), -(\cdot, \cdot), \times(\cdot, \cdot), \div(\cdot, \cdot), \text{sqrt}(\cdot),\text{power}(\cdot, \cdot), \log_e(\cdot), \exp(\cdot), \sin(\cdot), \cos(\cdot)\}.
\end{align*} 
This single layer OccamNet can be invoked by the LLM several times during generation to perform complex arithmetic operations accurately. To use the trained OccamLlama for inference, we sample the highest probability function from OccamNet as described in Appendix \ref{sec:probability}.

We benchmark our methods against unmodified Llama 2 7B Chat (Llama 2 7B)~\cite{llama2}, unmodified Llama 3 8B Instruct (Llama 3 8B)~\cite{llama3modelcard}, gpt-3.5-turbo-0125 (GPT 3.5 Turbo), gpt-4o-2024-05-13 (GPT 4o)~\cite{gpt4o}, and gpt-4o-2024-05-13 with Code Interpreter (GPT 4o + Code). To reduce costs, for GPT 4o with Code Interpreter, we test a random subset of 200 datapoints for each dataset.

To determine if a model output is correct, we parse all numbers in the model output and if one of them ``matches'' the correct answer, we determine that the result is correct. We mark each correct result as 100\% accuracy and each incorrect result as 0\% accuracy. For each model on each dataset, we report the mean accuracy and the standard error of the mean.
To determine if a number matches the result, we first determine how many places after the decimal $d$ the number should be accurate to. If the number is an integer, we set $d$ to 2. Otherwise, we set $d$ to the number of places after the decimal in the model output, clipped between 2 and 5. Finally we state that a number ``matches'' the result if the number and the result differ by less than $10^{-d}$.
We present further experiment details, including additional experiments, hyperparameters, and prompts in Appendix \ref{app: more experiments}.

\subsection{Simple Arithmetic Problems}\label{sec: simple arith}

\begin{table}[t]
\centering
\caption{Accuracy on arithmetic tasks, in percentages. The OccamLlama column corresponds to the results of both OccamLlama 8B and OccamLlama 70B. Higher is better. Bold indicates best performance for each row.}
\begin{tabular}{lcccccc}
\toprule
& OccamLlama & Llama 2 & Llama 3 & GPT 3.5 & GPT 4o & GPT 4o \\ 
& 8B / 70B & 7B Chat & 8b Instruct & Turbo &  & Code \\ 
\cmidrule(r){2-2}
\cmidrule(r){3-3}
\cmidrule(r){4-4}
\cmidrule(r){5-5}
\cmidrule(r){6-6}
\cmidrule(r){7-7}
Addition & \textbf{100.0\scriptsize{\textcolor{gray}{$\pm$0.0}}} & 19.2\scriptsize{\textcolor{gray}{$\pm$1.2}} & 44.9\scriptsize{\textcolor{gray}{$\pm$1.6}} & 65.2\scriptsize{\textcolor{gray}{$\pm$1.5}} & 95.7\scriptsize{\textcolor{gray}{$\pm$0.6}} & \textbf{100.0\scriptsize{\textcolor{gray}{$\pm$0.0}}} \\ 
Subtraction & \textbf{100.0\scriptsize{\textcolor{gray}{$\pm$0.0}}} & 8.7\scriptsize{\textcolor{gray}{$\pm$0.9}} & 34.4\scriptsize{\textcolor{gray}{$\pm$1.5}} & 59.8\scriptsize{\textcolor{gray}{$\pm$1.6}} & 85.6\scriptsize{\textcolor{gray}{$\pm$1.1}} & 99.5\scriptsize{\textcolor{gray}{$\pm$0.5}} \\ 
Multiplication & \textbf{100.0\scriptsize{\textcolor{gray}{$\pm$0.0}}} & 0.0\scriptsize{\textcolor{gray}{$\pm$0.0}} & 0.0\scriptsize{\textcolor{gray}{$\pm$0.0}} & 0.0\scriptsize{\textcolor{gray}{$\pm$0.0}} & 0.0\scriptsize{\textcolor{gray}{$\pm$0.0}} & 99.0\scriptsize{\textcolor{gray}{$\pm$0.7}} \\ 
Division & \textbf{100.0\scriptsize{\textcolor{gray}{$\pm$0.0}}} & 2.8\scriptsize{\textcolor{gray}{$\pm$0.5}} & 35.3\scriptsize{\textcolor{gray}{$\pm$1.5}} & 10.7\scriptsize{\textcolor{gray}{$\pm$1.0}} & 38.6\scriptsize{\textcolor{gray}{$\pm$1.5}} & \textbf{100.0\scriptsize{\textcolor{gray}{$\pm$0.0}}} \\ 
Square Root & \textbf{100.0\scriptsize{\textcolor{gray}{$\pm$0.0}}} & 0.0\scriptsize{\textcolor{gray}{$\pm$0.0}} & 0.0\scriptsize{\textcolor{gray}{$\pm$0.0}} & 0.9\scriptsize{\textcolor{gray}{$\pm$0.3}} & 18.6\scriptsize{\textcolor{gray}{$\pm$1.2}} & \textbf{100.0\scriptsize{\textcolor{gray}{$\pm$0.0}}} \\ 
Exponential & \textbf{100.0\scriptsize{\textcolor{gray}{$\pm$0.0}}} & 0.3\scriptsize{\textcolor{gray}{$\pm$0.2}} & 3.1\scriptsize{\textcolor{gray}{$\pm$0.5}} & 12.5\scriptsize{\textcolor{gray}{$\pm$1.0}} & 23.2\scriptsize{\textcolor{gray}{$\pm$1.3}} & \textbf{100.0\scriptsize{\textcolor{gray}{$\pm$0.0}}} \\ 
Logarithm & \textbf{100.0\scriptsize{\textcolor{gray}{$\pm$0.0}}} & 0.1\scriptsize{\textcolor{gray}{$\pm$0.1}} & 0.0\scriptsize{\textcolor{gray}{$\pm$0.0}} & 17.1\scriptsize{\textcolor{gray}{$\pm$1.2}} & 21.3\scriptsize{\textcolor{gray}{$\pm$1.3}} & \textbf{100.0\scriptsize{\textcolor{gray}{$\pm$0.0}}} \\ 
Sine & \textbf{100.0\scriptsize{\textcolor{gray}{$\pm$0.0}}} & 7.6\scriptsize{\textcolor{gray}{$\pm$0.8}} & 7.0\scriptsize{\textcolor{gray}{$\pm$0.8}} & 13.4\scriptsize{\textcolor{gray}{$\pm$1.1}} & 39.3\scriptsize{\textcolor{gray}{$\pm$1.5}} & \textbf{100.0\scriptsize{\textcolor{gray}{$\pm$0.0}}} \\ 
Cosine & \textbf{100.0\scriptsize{\textcolor{gray}{$\pm$0.0}}} & 0.8\scriptsize{\textcolor{gray}{$\pm$0.3}} & 1.5\scriptsize{\textcolor{gray}{$\pm$0.4}} & 6.7\scriptsize{\textcolor{gray}{$\pm$0.8}} & 32.8\scriptsize{\textcolor{gray}{$\pm$1.5}} & \textbf{100.0\scriptsize{\textcolor{gray}{$\pm$0.0}}} \\ 
\midrule
\textsc{Average} & \textbf{100.0\scriptsize{\textcolor{gray}{$\pm$0.0}}}& 4.4\scriptsize{\textcolor{gray}{$\pm$0.2}} & 14.0\scriptsize{\textcolor{gray}{$\pm$0.4}} & 20.7\scriptsize{\textcolor{gray}{$\pm$0.4}} & 39.5\scriptsize{\textcolor{gray}{$\pm$0.5}} & 99.8\scriptsize{\textcolor{gray}{$\pm$0.1}} \\

\bottomrule
\end{tabular}
\label{tab:arithmetic_accuracy}
\end{table}

To evaluate OccamLlama and the baselines on purely arithmetic expressions, we create several synthetic datasets. For each of the operations in $\{+,-,\times,\div\}$, the inputs are random 7-digit positive or negative integers. For $\sqrt{}$, the inputs are random 7-digit positive integers. For the logarithms, the examples are log-uniformly sampled in the interval $(10^{-10}, 10^{10})$; for the exponentials, they are uniformly sampled in the interval $(-10, 10)$, and for sines and cosines they are uniformly sampled in the interval $(-2\pi,2\pi)$.

The results of these evaluations are shown in Table \ref{tab:arithmetic_accuracy}. More detailed results, including relative error and results for 3- and 5-digit arithmetic, are shown in Appendix \ref{app: exp results}. 

Both OccamLlama 8B and 70B have $100.0 \pm 0.0$\% accuracy on all tasks, missing 0 out of 9000 problems. On the other hand, we tested GPT 4o with Code Interpreter on fewer problems to save cost, and it missed 3 out of the 1800 problems it faced, achieving an accuracy of $99.8 \pm 0.1$\%.

Furthermore, GPT 4o with Code Interpreter generates on average more than 54 tokens to answer these problems, whereas our model uses OccamNet on the first forward pass. This means that, barring advanced decoding techniques such as speculative decoding \cite{spec_decoding}, GPT 4o would need to be more than 50x faster than OccamLlama per forward pass to be comparable in answer generation speed on these tasks.

\begin{figure}
    \centering
    \includegraphics[width=\textwidth]{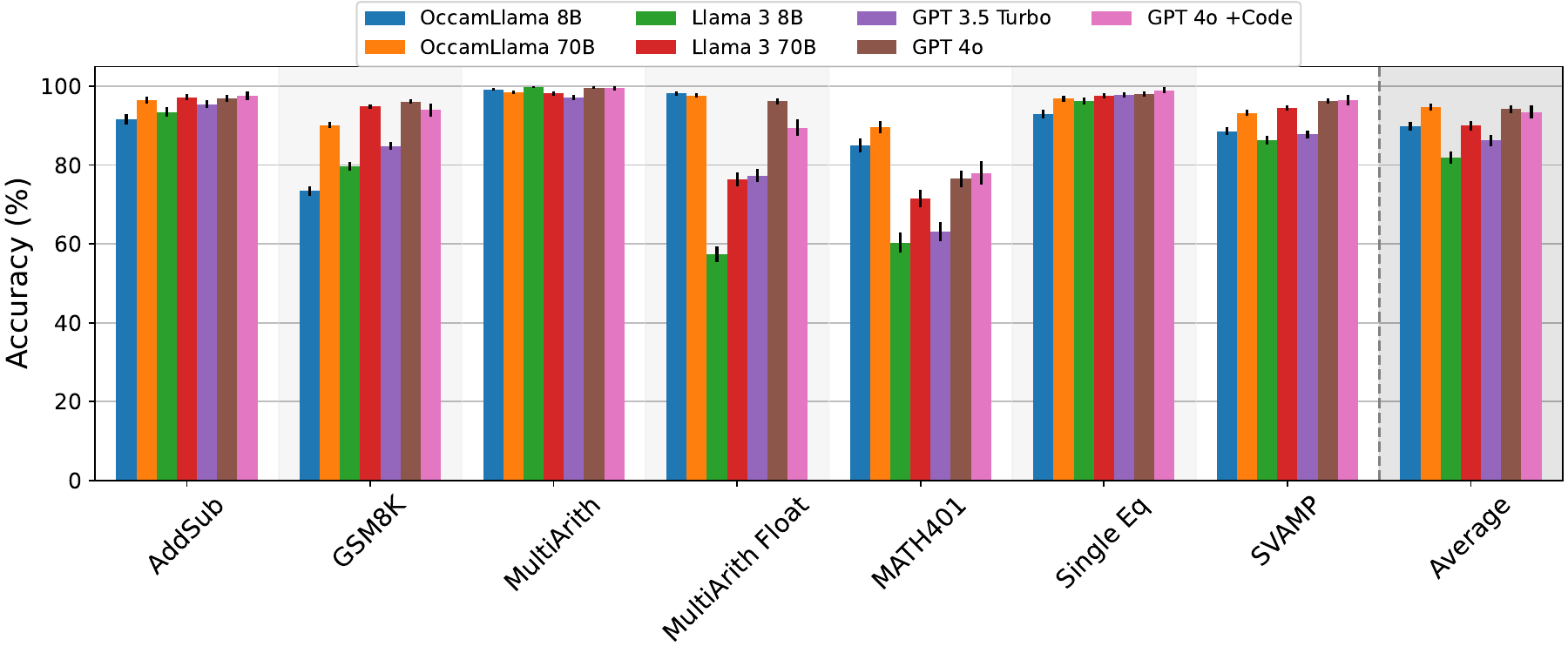}
    \vspace{-2em}
    \caption{Accuracy of OccamLlama and baselines on mathematical problem solving tasks. Higher is better. OccamLlama 8B achieves accuracy comparable to Llama 3 8B on benchmarks with simple arithmetic, higher accuracy than GPT 4o and GPT 4o + Code on on tasks with challenging arithmetic, and accuracy above Llama 3 8B and similar to GPT 3.5 Turbo on average. OccamLlama 70B outperforms GPT 4o and GPT 4o + Code on average.}
    \vspace{-1em}
    \label{fig:reasoning}
\end{figure}

Table \ref{tab:arithmetic_accuracy} demonstrates that arithmetic with LLMs is still challenging; state-of-the-art proprietary language models like GPT 4o achieve less than 40\% accuracy on 7-digit division and fail to perform any 7-digit multiplications correctly. Open source LLMs fall farther behind, with Llama 3 8B achieving below 50\% on relatively simple tasks such as 7-digit addition.

\subsection{Mathematical Problem Solving}\label{sec: reasoning}

To test the performance of OccamLlama on more general mathematical problem solving tasks, we evaluate our method and baselines on the following six benchmarks: AddSub \cite{addsub}, GSM8K \cite{gsm8k}, MultiArith \cite{multiarith}, MATH401 \cite{math401}, Single Eq \cite{singleeq}, and SVAMP \cite{svamp}. All but MATH401 are word problems requiring longer generation and a mix of reasoning and arithmetic capabilities. MATH401 also includes multistep arithmetic problems which require more than one call to OccamLlama. We selected these datasets (including the MultiArith Float dataset described below) before testing any methods on them to ensure unbiased selection of benchmarks.

Because many of the arithmetic operations required in these datasets are relatively simple, we also create MultiArith Float, a modification of MultiArith in which we select problems which are arithmetically more challenging, while requiring similar levels of reasoning. To this end, we select prompts having input numbers that can be replaced with floats. For instance, $3.5$ feet or $\$39.95$ are reasonable but $3.5$ people is not. Furthermore, we sample input values from ranges larger than those appearing in the MultiArith dataset, in cases where it is reasonable. Float operations and larger additions and multiplications are more difficult for the baseline LLMs but do not make a difference for OccamLLM, so this dataset is particularly useful to show the advantages of the system we propose. Figure \ref{fig:reasoning} shows the results of these evaluations. More detailed results are shown in Appendix \ref{app: exp results}. 

OccamLlama 70B outperforms both GPT 4o and GPT 4o + Code on average across the benchmarks, demonstrating OccamLlama's strong mathematical problem solving capability. We also note that GPT 4o + Code does not outperform GPT 4o on average, suggesting that existing implementations of LLMs with code generation may not help with mathematical problem solving.

We now consider the performance of OccamLlama 8B, the smaller OccamLlama model. On MultiArith Float and MATH401, two datasets requiring challenging arithmetic, OccamLlama 8B outperforms not only Llama 3 8B but also GPT 4o and GPT 4o + Code.
At the same time, most other datasets in this benchmark do not involve challenging arithmetic, meaning that Llama 3 8B is well suited to solve these tasks without assistance; most of the difficulty of these tasks lies in the reasoning rather than in the arithmetic computations. This is further supported by the fact that GPT 4o with Code Interpreter never substantially outperforms and sometimes underperforms GPT 4o on these tasks. As such, it is remarkable that OccamLlama 8B can achieve comparable accuracy to Llama 3 8B even when it is trained on very different data and evaluated on tasks without challenging arithmetic. 

The only datasets for which OccamLlama 8B performs noticeably worse than Llama 3 8B are GSM8K and Single Eq, but we believe this results from an imperfect OccamLlama switch, likely stemming from text which is outside of the switch training distribution (see Section \ref{sec:limitations}). Fortunately, in Appendix \ref{app: robustness}, we find that the OccamNet decoder is quite robust to out of distribution data and that both the OccamNet and switch decoders generalize well to unseen languages. This suggests that, with relatively little data, it should be possible to teach the switch to handle these unseen cases, something we leave for future work.

\begin{figure}
    \centering
    \includegraphics[width=\textwidth]{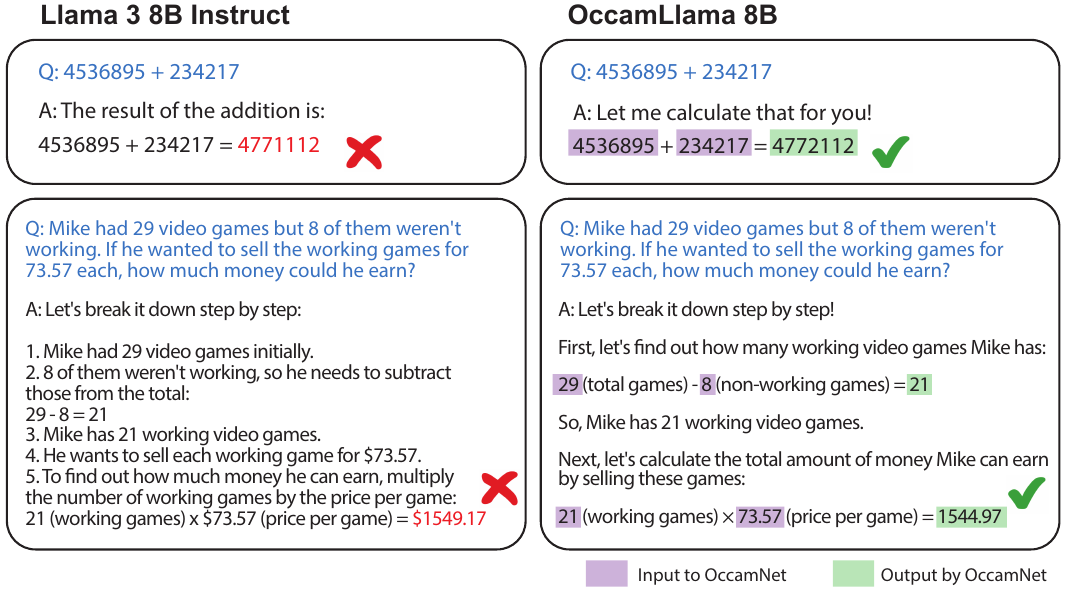}
    \caption{Examples from Llama 3 8B Instruct and OccamLlama 8B on (top) an arithmetic problem and (bottom) a mathematical reasoning problem from the MultiArith Float dataset. In OccamLlama, the LLM performs reasoning, the switch predicts when to use OccamNet, and OccamNet performs arithmetic operations. OccamNet's inputs and outputs are highlighted in purple and green, respectively.}
    \label{fig: example_generations}
\end{figure}

In Figure \ref{fig: example_generations}, we show example generations from OccamLlama 8B for both arithmetic and reasoning tasks. These generations demonstrate how the OccamLlama switch learns to balance OccamNet outputs with LLM outputs, effectively distributing the work between a reasoner (Llama) and a calculator (OccamNet). Because the language model is unaware of the OccamLlama system, its generations behave as if it possesses an interior calculator even though it is actually using a tool. In this way, we combine the benefits of a language model finetuned on arithmetic with the benefits of a language model finetuned to use code for arithmetic, all \textit{without any finetuning}.

\subsection{Limitations}\label{sec:limitations}
In our experiments, we use a single-layer OccamNet as the symbolic network, enabling evaluation of \textit{single-operation} arithmetic problems. This sometimes poses a challenge on reasoning problems when the base language model generates compound expressions requiring more than one operation to evaluate, such as $3+5+7 = $. A single-layer OccamNet cannot evaluate these expressions. We attempted to overcome this by prompting Llama to break down compound expressions into multiple steps, but we find it difficult to coerce Llama to follow these instructions. Another challenge is that Llama often generates expressions in fractions or percentages, which also constitute compound expressions that are not properly handled by the OccamLLM system. Fortunately, we observed that these compound expressions were typically simple enough for the LLM to evaluate without OccamNet. Therefore, in our experiments, we trained the OccamLLM switch to avoid using OccamNet for compound operations, largely mitigating this issue. Future work could explore other solutions such as integrating a two-layer OccamNet as the symbolic network. We found that these issues are particularly acute in the GSM8K and Single Eq datasets, where the expressions generated by Llama are not prevalent in the switch training data, causing it to sometimes incorrectly trigger OccamNet and degrade performance, as discussed more in Appendix \ref{app: exp results}.

Furthermore, we found that the language model sometimes appends further digits to OccamLlama outputs, defeating the purpose of OccamLlama generations. To address this issue, we append ``\textbackslash n\textbackslash n.'' to every number computed with OccamNet, emulating the usual behavior of Llama. 

These techniques demonstrate a design paradigm of OccamLlama: by tuning the behaviors of OccamNet and the switch, we can often avoid finetuning the LLM.
\looseness-1

\section{Discussion}

We presented OccamLLM, a system enabling exact and interpretable language model arithmetic in a single autoregressive step. Our method does not require modifying the weights of the underlying language model, thereby avoiding risks of catastrophic forgetting. Furthermore, our method avoids security risks arising from running code generated by a language model while outperforming top LLM code generation methods (GPT 4o + Code) on average across our benchmarks.

We benchmarked our method on challenging arithmetic tasks, achieving 100\% accuracy where GPT 4o achieves only 40\% performance on average. We also benchmarked our method on mathematical problem solving tasks, demonstrating that the OccamLlama switch can accurately balance the LLM for reasoning and OccamNet for arithmetic, outperforming even GPT 4o and GPT 4o with Code Interpreter on average.

Our work could enable smaller LLMs to be as performant as much larger LLMs in arithmetic. Moreover, integrating OccamLLM with larger LLMs like GPT 4o could further improve their arithmetic abilities without requiring a code interpreter.
Furthermore, at present, OccamLLM may not integrate with more advanced decoding techniques such as speculative decoding \cite{spec_decoding, staged_spec_decoding}. We hope to explore these avenues in future work.

\section{Broader Impact}\label{sec: impact}

We believe that, in addition to enabling fast, safe, and interpretable arithmetic, OccamLLM demonstrates a new paradigm for tool use. As a proof of concept for more complex tool use, we further train OccamLlama 8B with a two layer Complete OccamNet with the primitives
\begin{equation*}
    \mathcal{P} = \{\mathsf{Addition}(\cdot, \cdot), \mathsf{Subtraction}(\cdot, \cdot), \mathsf{Multiplication}(\cdot, \cdot), \mathsf{Division}(\cdot, \cdot)\},
\end{equation*}
which enables OccamLlama to perform up to three arithmetic operations (e.g., $2\cdot 7+3/2$) in a single autoregressive step. We find that this two-layer OccamLlama can reach near 100\% accuracy, even when performing three arithmetic operations in a single autoregressive step, as shown in Table \ref{tab: multi_layer_occam}. This demonstrates that OccamLLM can be used to perform more complex operations, including composing multiple different tools.

\begin{wraptable}{r}{0.45\textwidth} 
\centering
\caption{Accuracy on multistep arithmetic.}
\begin{tabular}{lcc}
\toprule
& OccamLlama & Llama 3 \\ 
&  & 8b Instruct \\ 
\cmidrule(r){2-2}
\cmidrule(r){3-3}
One-Step & \textbf{99.9\scriptsize{\textcolor{gray}{$\pm$0.1}}} & 78.1\scriptsize{\textcolor{gray}{$\pm$1.3}} \\ 
Two-Step & \textbf{98.2\scriptsize{\textcolor{gray}{$\pm$0.4}}} & 57.8\scriptsize{\textcolor{gray}{$\pm$1.6}} \\ 
Three-Step & \textbf{96.1\scriptsize{\textcolor{gray}{$\pm$0.6}}} & 40.2\scriptsize{\textcolor{gray}{$\pm$1.6}} \\ 
\midrule
\textsc{Average} & \textbf{98.1\scriptsize{\textcolor{gray}{$\pm$0.3}}}& 58.7\scriptsize{\textcolor{gray}{$\pm$0.9}} \\

\bottomrule
\end{tabular}
\label{tab: multi_layer_occam}
\end{wraptable}

For future work, we plan to explore integrating other tools beyond calculators through a similar technique. This is facilitated by the fact that there are no restrictions on OccamNet's activations; in principle, tools could be placed inside activations of OccamNet, enabling OccamNet to serve as a sort of a mixture of experts for tools. 
While some tools, like querying a search engine, may still be most effective when integrated into language model systems through language, we believe this work demonstrates that some tools are more effective when they can be more tightly integrated into the language model.
\looseness-1

\section*{Acknowledgements}
We would like to thank Andrew Ma and Di Luo for their thoughtful discussions.

The authors acknowledge the MIT SuperCloud and Lincoln Laboratory Supercomputing Center for providing HPC resources that have contributed to the research results reported within this paper.

Research was sponsored by the Department of the Air Force Artificial Intelligence Accelerator and was accomplished under Cooperative Agreement Number FA8750-19-2-1000. The views and conclusions contained in this document are those of the authors and should not be interpreted as representing the official policies, either expressed or implied, of the Department of the Air Force or the U.S. Government. The U.S. Government is authorized to reproduce and distribute reprints for Government purposes notwithstanding any copyright notation herein.

This work is also supported in part by the National Science Foundation under Cooperative Agreement PHY-2019786
(The NSF AI Institute for Artificial Intelligence and Fundamental Interactions, \url{http://iaifi.org/}).

\bibliographystyle{unsrt}
\bibliography{ref}


\input{appendix}


\end{document}

%% file: appendix.tex
\newpage
\section*{Appendix}
\appendix

\section{Further Experiment Details and Results}\label{app: more experiments}

For training and evaluation OccamLlama 8B, we used a single 32 GB NVIDIA Tesla V100 GPU. For OccamLlama 70B, we used two 80 GB NVIDIA A100 GPU. 

In the experiments presented in Section \ref{sec: experiments}, for each of the weight decoders and the switch, we used two-layer MLPs of input size 4096/8192 (Llama 3 8B/70B Instruct hidden size), intermediate size 64 and final size equal to the number of weights in the corresponding OccamNet layer or switch. 

In the two-layer experiments presented in Section \ref{sec: impact}, for each of the weight decoders, we used two-layer MLPs of input size 4096 (Llama 3 8B Instruct hidden size), intermediate size 512, and final size equal to the number of weights in the corresponding OccamNet layer. We did not train a switch for this experiment as we did not test long-form generations.

\subsection{Training Dataset} \label{app: dataset}

\subsubsection{OccamNet Decoder} \label{app: dataset occamnet}
To train the OccamNet decoder, we created a training dataset consisting of a 80,000 examples split in 40,000 single queries and 40,000 sequences of concatenated queries. In the first case, we sampled a single prompt of those described in \ref{sec: Data Generation} and formatted it using the Llama 3 Instruct chat template. In the second case, we concatenated multiple prompts described in \ref{sec: Data Generation} without the chat template.

40\% of the sampled prompts correspond to simple arithmetic, concretely $+,-,\times$, and $\div$. We sampled from various input value ranges, chosen at random: integers in $[-10,10]$, integers in $[-100,100]$, integers in $[-1000,1000]$, integers in $[-20000, 20000]$, floating numbers in $[-1,1]$, and floating point numbers in $[-1000,1000]$.

Another 40\% corresponds to complex arithmetic involving square roots, logarithms, exponentials, trigonometric functions and computing one number to the power of another. For the square root and the logarithm, we sampled integers uniformly in either $[1, 100]$ or $[1,20000]$ and floats uniformly in either $[0.01, 100]$ or $[0.01,20000]$. For the exponential, we sampled integers and floats in $[-10,10]$. For the powers, we sampled the base as either an integer in $[1,25]$ or a float in $[0.1,25]$ and the exponent as an integer in $[-6,6]$.

The remaining 20\% corresponds to single or multi step problems reasoning prompts. The inputs were sampled with various ranges, sometimes as floats and sometimes as integers, depending on the context of the problem. Because a single-OccamNet-layer OccamLlama cannot solve a multi-step reasoning problem in a single step, we never end the multiple-query examples with a multi-step reasoning problem.

We first iterated the 80,000 examples, prepending ``Answer = '' to the assistant response, thus training OccamNet to predict the result after the ``=''. Next, we validated the model on out-of-distribution examples where ``Answer = '' was not appended. We noticed that the accuracy on this task was improving during training, but after the full dataset was iterated it still didn't perform as well as when evaluated in-distribution. Therefore, we continued to train the model using examples of the same dataset but with no ``Answer = '' at the beginning of the assistant response. The model rapidly learned the new task. We stopped at 28,000 iterations of this second stage.

For the two-layer OccamNet run, we generated a large set of programmatically generated prompts of the form $3+97\cdot -4 =$, with the Llama 3 Instruct chat template applied.

\subsubsection{Switch Decoder}  \label{app: dataset switch}
To train the switch decoder, we created a dataset of 50,000 examples (80,000 for OccamLlama 80B). For each example, the tokens previous to the numbers that should be computed using OccamNet, which are the ones that the switch should not route to the LLM, are labeled with a 1, and all the rest are labeled with a 0. 

Half of the examples consist of a single prompt corresponding to a simple arithmetic expression as the ones described in Section \ref{sec: Data Generation}. The token immediately at the beginning of the assistant response is labeled with a 1. Therefore, the trained system will answer directly to simple arithmetic queries that OccamNet can compute. 

The remaining 25,000 examples consist each of a series of prompts which are formatted in the Llama 3 Instruct chat template in a conversational style. The input-output pairs used to create each sequence of prompts are distributed in the following way:

\begin{itemize}
    \item 25\% of these pairs are created by taking one of the simple arithmetic expressions as input. The output is selected randomly between answering directly at the beginning of the assistant response, adding "Answer = " before the answer, or repeating the input expression before the answer. These examples train the switch to trigger OccamNet in different scenarios where the LLM needs to compute an answer.

    \item 70\% of the pairs come from a collection of 43 manually created and labeled examples, which illustrate in which cases the switch should route to OccamNet and, importantly, in which cases it shouldn't. This collection was designed to cover a wide variety of situations where the LLM might need to use OccamNet for computations. Furthermore, it includes cases where the LLM should avoid calling OccamNet because doing so would produce a bad prediction. This is the case, for example, of instances where the LLM attempts to add three numbers simultaneously. If it were to use the 1-layer OccamNet, which can take 2 inputs at most, the result would be incorrect. 

    \item The remaining 5\% of the prompts come from multi-step reasoning problems. We set the output for these not to a full response, but only ``The answer is ''. In such cases, a single-layer OccamNet cannot compute the answer, so the output tokens are labeled with a 0. This trains the system to avoid routing to OccamNet when the later cannot compute the answer.

\end{itemize}

Below, we present some manually created examples and explain what case they cover. The tokens of the numbers in \textbf{bold} were marked with a 1 to signal that the switch should route to OccamNet, and the rest of the numbers and all of the text tokens were marked with a 0 to signal that OccamNet should not be called in these cases.

The cases we cover with these prompts are:
\begin{enumerate}
    \item Single-layer OccamNet should be called to perform two-input arithmetic operations.
    \item OccamNet should not be called to write numbers that already appeared and do not require further arithmetic.
    \item OccamNet should not be called to round numbers.
    \item OccamNet should not be called to perform inline operations nested within a larger expression because it might contradict the assumption that the last two numbers are involved in the operation.
    \item OccamNet should not be used to convert magnitudes when the LLM does not explicitly write the conversion factor and relies on its internal knowledge; without the numeric value of the factor present in the text, OccamNet can't give the correct result.
    \item OccamNet can't compute percentages, as this operation was not included in the training set of operations. It could be done in two steps.
    \item Single-layer OccamNet can't compute a fraction of a number, as this would require two operations.
    \item Single-layer OccamNet can't perform operations with more than 2 arguments.
\end{enumerate}

\renewcommand{\arraystretch}{1.9}
\begin{longtable}{|p{5cm}|p{7cm}|c|}
\hline
\textbf{User Query} & \textbf{Assistant Response} & \textbf{Cases} \\ 
\hline
\endfirsthead
\hline
\textbf{User Query} & \textbf{Assistant Response} & \textbf{Cases} \\ 
\hline
\endhead
\hline
\endfoot
\hline
\endlastfoot
An author writes 7.6 pages per session. After 9 sessions, the total pages written are & 
The author writes 7.6 pages per session. After 9 sessions, the author will have written 7.6 \texttimes 9 = \textbf{68.4} pages. The answer is 68. & 1, 2, 3 \\ \hline

I have 10 oranges and 6 apple trees, each of them with 3 apples. How much fruit do I have? & 
The total number of pieces of fruit is 10 oranges + 6 trees times 3 apples = 10 + (6 \texttimes 3) = 10 + 18 = \textbf{28}. The answer is 28. & 1, 2, 4 \\ \hline

Sally has 7 pennies and 8 nickels. How many pennies does she have in total? & 
1. Convert the number of nickels to pennies: 8 nickels is 40 pennies. \newline 
2. Add total number of pennies 40 + 7 = \textbf{47}. \newline 
The answer is 47. & 1, 2, 5 \\ \hline

What is 84 percent of 230.77? & 
84\% of 230.77 = 193.85 \newline
Explanation: \newline
84 / 100 = \textbf{0.84} \newline
0.84 \texttimes 230.77 = \textbf{193.85} & 1, 2, 6 \\ \hline

There were 39 cats on a rock. 3 boats came and carried away 5 cats each. How many cats were left? & 
There were originally 39 cats. 3 boats came and each took away 5 cats. So 3 \texttimes 5 = \textbf{15}. \newline 
39 - 15 = \textbf{24}. \newline
Then 3/7 of them ran away. 3/7 of 24 is 9. \newline
24 - 9 = \textbf{15}. \newline
So there were 15 cats left. & 1, 2, 7 \\ \hline

11 people have 2 apples, 2 oranges, and 2 peaches each. How many pieces of fruit do they have? & 
Each person has 2 + 2 + 2 = 6 pieces of fruit. \newline
In total, they have 6 \texttimes 11 = \textbf{66}. \newline
The answer is 66. & 1, 2, 8 \\ \hline
\end{longtable}

\subsection{Training Hyperparameters}

For all 1-layer OccamNet training runs, we used a batch size of $1$, a learning rate of $6e-4$ and a weight decay parameter of $0.01$. We use gradient accumulation to achieve an effective batch size of $8$. We used a constant learning rate scheduler. We take 1000 samples from OccamNet per token.

For the 2-layer OccamNet run, we used a batch size of $1$, a learning rate of $1e-4$ and a weight decay parameter of $0.01$. We use the gradient accumulation technique to achieve an effective batch size of $8$. We used a constant learning rate scheduler. We take 50,000 samples from OccamNet per token.

\subsection{Prompting}
For the division arithmetic tasks, we found that the language models often did not return decimals. As such, we appended ``Give the answer in decimals.'' to these prompts. Similarly, for the trigonometric functions evaluations, we explicitly ask the language models to take the input as radians, by formatting the prompts as "cos(X rad) =".

For some models, we provide system prompting to guide the model toward the correct behavior. We break down prompting by model below:

\paragraph{Llama 2/3:} We did not provide a system prompt for the arithmetic tasks. For the reasoning tasks, we used the system prompt ``Solve step by step.''

\paragraph{GPT 3.5 Turbo:} We do not use a system prompt for GPT 3.5 Turbo.

\paragraph{GPT 4o:} We did not use a system prompt, except for the MATH401 dataset, where we noticed that GPT 4o was returning fractions instead of decimals. As such, on MATH401 we used the system prompt ``Give your answer in decimals.'' 

\paragraph{GPT 4o + Code:} We used the system prompt ``Write and run code to answer math questions. Do not format numbers. Give all answers in decimals.''

\paragraph{OccamLlama:} We experimented with OccamLlama prompts, but discovered that not including a system prompt was most effective.

\subsection{Generation parameters}

For OccamLlama, Llama 2 7B and Llama 3 8B, we use the default values of $T=0.6$ and Top-P $= 0.9$. For GPT 3.5 Turbo, GPT 4o, and GPT 4o with Code Interpreter, we use the default values of $T=1.0$ and Top-P $= 1.0$.

\subsection{Experimental Results}\label{app: exp results}

Tables \ref{tab:arithmetic_accuracy_app} and \ref{tab:reasoning_accuracy} show in more detail the accuracy of OccamLlama and other baselines on arithmetic and mathematical problem solving tasks. We measure accuracy as described in the main text.

We note here that on datasets with challenging arithmetic, in particular Multiarith Float and MATH401, OccamLlama 8B outperforms even GPT 4o and GPT 4o Code. In fact, on MultiArith Float, OccamLlama 8B is nearly 10 percentage points more accurate than GPT 4o + Code and and more than 40 percentage points more accurate than Llama 3 8B. Similarly, on MATH401, OccamLlama 8B is 7 percentage points more accurate than GPT 4o + Code and nearly 25 percentage points more accurate than Llama 3 8B. Although MATH401 does not include word problems, it does include some arithmetic expressions that require multiple calls to OccamNet to solve, meaning it requires both reasoning (to determine how to break up the arithmetic expression) and arithmetic capabilities.

The only datasets on which OccamLlama 8B performs substantially worse than Llama 3 8B are GSM8K \cite{gsm8k} and Single Eq \cite{singleeq}. We believe a contributor to this is that these datasets include many problems that involve either fractions and percentages, which Llama does not convert to decimal format, or equations with unknown variables. As such, Llama often calls OccamNet with expressions such as ``multiplying by 3/4 gives,'' ``5\% of this gives,'' or ``adding 5 to both sides of x-5 = 11 gives.'' Because the switch is not trained on many examples like these in which the number is not in decimal format, it does not realize that OccamNet should not be used in these cases. Therefore, the switch triggers OccamNet, which is not capable of performing the correct operation (these types of operations are not acheivable with a 1-layer OccamNet). Future work could address this issue by training the switch with more data on this type of situation or by training an OccamLlama with a two layer OccamNet. 

Finally, as noted in the main text, OccamLlama 70B achieves significant performance improvement over OccamLlama 8B across a number of benchmarks and outperforms GPT 4o and GPT 4o + Code on average. This demonstrates that OccamLLM improves with the base language model and suggests that combining OccamLLM with more capable models such as GPT 4o could be a promising avenue for future research.

Relative error is another important metric that complements accuracy. It measures by how much the answer differs from the true result. For two models with a similar accuracy metric, the relative error they achieve can be very different. Table \ref{tab:arithmetic_error_app} shows the relative error for the arithmetic experiments. An answer marked correct can have a nonzero relative error because of machine precision limits and because the answer does not report an infinite number of digits.

Interestingly, Llama 2 performs exceptionally poorly on division. By examining outputs, we see that this is because Llama 2 produces an approximately correct output but with the decimal place in the wrong position, leading to a result that is off by many orders of magnitude.

\begin{table}[t]
\centering
\caption{Percent accuracy on arithmetic tasks. Higher is Better. Bold indicates best performance.}
\begin{tabular}{lcccccc}
\toprule
& OccamLlama & Llama 2 & Llama 3 & GPT 3.5 & GPT 4o & GPT 4o \\ 
&  & 7B Chat & 8b Instruct & Turbo &  & Code \\ 
\cmidrule(r){2-2}
\cmidrule(r){3-3}
\cmidrule(r){4-4}
\cmidrule(r){5-5}
\cmidrule(r){6-6}
\cmidrule(r){7-7}
Addition (3) & \textbf{100.0\scriptsize{\textcolor{gray}{$\pm$0.0}}} & 70.9\scriptsize{\textcolor{gray}{$\pm$1.4}} & 97.1\scriptsize{\textcolor{gray}{$\pm$0.5}} & 98.8\scriptsize{\textcolor{gray}{$\pm$0.3}} & \textbf{100.0\scriptsize{\textcolor{gray}{$\pm$0.0}}} &  \\ 
Addition (5) & \textbf{100.0\scriptsize{\textcolor{gray}{$\pm$0.0}}} & 55.9\scriptsize{\textcolor{gray}{$\pm$1.6}} & 77.1\scriptsize{\textcolor{gray}{$\pm$1.3}} & 92.5\scriptsize{\textcolor{gray}{$\pm$0.8}} & 99.2\scriptsize{\textcolor{gray}{$\pm$0.3}} &  \\ 
Addition (7) & \textbf{100.0\scriptsize{\textcolor{gray}{$\pm$0.0}}} & 19.2\scriptsize{\textcolor{gray}{$\pm$1.2}} & 44.9\scriptsize{\textcolor{gray}{$\pm$1.6}} & 65.2\scriptsize{\textcolor{gray}{$\pm$1.5}} & 95.7\scriptsize{\textcolor{gray}{$\pm$0.6}} & \textbf{100.0\scriptsize{\textcolor{gray}{$\pm$0.0}}} \\ 
\\
Subtraction (3) & \textbf{100.0\scriptsize{\textcolor{gray}{$\pm$0.0}}} & 49.7\scriptsize{\textcolor{gray}{$\pm$1.6}} & 95.2\scriptsize{\textcolor{gray}{$\pm$0.7}} & 94.0\scriptsize{\textcolor{gray}{$\pm$0.8}} & 98.7\scriptsize{\textcolor{gray}{$\pm$0.4}} &  \\ 
Subtraction (5) & \textbf{100.0\scriptsize{\textcolor{gray}{$\pm$0.0}}} & 22.9\scriptsize{\textcolor{gray}{$\pm$1.3}} & 58.8\scriptsize{\textcolor{gray}{$\pm$1.6}} & 86.3\scriptsize{\textcolor{gray}{$\pm$1.1}} & 92.6\scriptsize{\textcolor{gray}{$\pm$0.8}} &  \\ 
Subtraction (7) & \textbf{100.0\scriptsize{\textcolor{gray}{$\pm$0.0}}} & 8.7\scriptsize{\textcolor{gray}{$\pm$0.9}} & 34.4\scriptsize{\textcolor{gray}{$\pm$1.5}} & 59.8\scriptsize{\textcolor{gray}{$\pm$1.6}} & 85.6\scriptsize{\textcolor{gray}{$\pm$1.1}} & 99.5\scriptsize{\textcolor{gray}{$\pm$0.5}} \\ 
\\
Multiplication (3) & \textbf{100.0\scriptsize{\textcolor{gray}{$\pm$0.0}}} & 4.6\scriptsize{\textcolor{gray}{$\pm$0.7}} & 16.8\scriptsize{\textcolor{gray}{$\pm$1.2}} & 49.2\scriptsize{\textcolor{gray}{$\pm$1.6}} & 76.9\scriptsize{\textcolor{gray}{$\pm$1.3}} &  \\ 
Multiplication (5) & \textbf{100.0\scriptsize{\textcolor{gray}{$\pm$0.0}}} & 0.0\scriptsize{\textcolor{gray}{$\pm$0.0}} & 0.1\scriptsize{\textcolor{gray}{$\pm$0.1}} & 0.4\scriptsize{\textcolor{gray}{$\pm$0.2}} & 4.6\scriptsize{\textcolor{gray}{$\pm$0.7}} &  \\ 
Multiplication (7) & \textbf{100.0\scriptsize{\textcolor{gray}{$\pm$0.0}}} & 0.0\scriptsize{\textcolor{gray}{$\pm$0.0}} & 0.0\scriptsize{\textcolor{gray}{$\pm$0.0}} & 0.0\scriptsize{\textcolor{gray}{$\pm$0.0}} & 0.0\scriptsize{\textcolor{gray}{$\pm$0.0}} & 99.0\scriptsize{\textcolor{gray}{$\pm$0.7}} \\ 
\\
Division (3) & \textbf{100.0\scriptsize{\textcolor{gray}{$\pm$0.0}}} & 20.8\scriptsize{\textcolor{gray}{$\pm$1.3}} & 71.7\scriptsize{\textcolor{gray}{$\pm$1.4}} & 50.5\scriptsize{\textcolor{gray}{$\pm$1.6}} & 78.2\scriptsize{\textcolor{gray}{$\pm$1.3}} &  \\ 
Division (5) & \textbf{100.0\scriptsize{\textcolor{gray}{$\pm$0.0}}} & 7.4\scriptsize{\textcolor{gray}{$\pm$0.8}} & 48.1\scriptsize{\textcolor{gray}{$\pm$1.6}} & 15.7\scriptsize{\textcolor{gray}{$\pm$1.2}} & 51.0\scriptsize{\textcolor{gray}{$\pm$1.6}} &  \\ 
Division (7) & \textbf{100.0\scriptsize{\textcolor{gray}{$\pm$0.0}}} & 2.8\scriptsize{\textcolor{gray}{$\pm$0.5}} & 35.3\scriptsize{\textcolor{gray}{$\pm$1.5}} & 10.7\scriptsize{\textcolor{gray}{$\pm$1.0}} & 38.6\scriptsize{\textcolor{gray}{$\pm$1.5}} & \textbf{100.0\scriptsize{\textcolor{gray}{$\pm$0.0}}} \\
\\
Square Root (3) & \textbf{100.0\scriptsize{\textcolor{gray}{$\pm$0.0}}} & 1.2\scriptsize{\textcolor{gray}{$\pm$0.3}} & 14.8\scriptsize{\textcolor{gray}{$\pm$1.1}} & 47.1\scriptsize{\textcolor{gray}{$\pm$1.6}} & 69.3\scriptsize{\textcolor{gray}{$\pm$1.5}} &  \\ 
Square Root (5) & \textbf{100.0\scriptsize{\textcolor{gray}{$\pm$0.0}}} & 0.2\scriptsize{\textcolor{gray}{$\pm$0.1}} & 1.3\scriptsize{\textcolor{gray}{$\pm$0.4}} & 11.9\scriptsize{\textcolor{gray}{$\pm$1.0}} & 23.6\scriptsize{\textcolor{gray}{$\pm$1.3}} &  \\ 
Square Root (7) & \textbf{100.0\scriptsize{\textcolor{gray}{$\pm$0.0}}} & 0.0\scriptsize{\textcolor{gray}{$\pm$0.0}} & 0.0\scriptsize{\textcolor{gray}{$\pm$0.0}} & 0.9\scriptsize{\textcolor{gray}{$\pm$0.3}} & 18.6\scriptsize{\textcolor{gray}{$\pm$1.2}} & \textbf{100.0\scriptsize{\textcolor{gray}{$\pm$0.0}}} \\ 
\\
Exponential & \textbf{100.0\scriptsize{\textcolor{gray}{$\pm$0.0}}} & 0.3\scriptsize{\textcolor{gray}{$\pm$0.2}} & 3.1\scriptsize{\textcolor{gray}{$\pm$0.5}} & 12.5\scriptsize{\textcolor{gray}{$\pm$1.0}} & 23.2\scriptsize{\textcolor{gray}{$\pm$1.3}} & \textbf{100.0\scriptsize{\textcolor{gray}{$\pm$0.0}}} \\ 
Logarithm & \textbf{100.0\scriptsize{\textcolor{gray}{$\pm$0.0}}} & 0.1\scriptsize{\textcolor{gray}{$\pm$0.1}} & 0.0\scriptsize{\textcolor{gray}{$\pm$0.0}} & 17.1\scriptsize{\textcolor{gray}{$\pm$1.2}} & 21.3\scriptsize{\textcolor{gray}{$\pm$1.3}} & \textbf{100.0\scriptsize{\textcolor{gray}{$\pm$0.0}}} \\ 
Sine & \textbf{100.0\scriptsize{\textcolor{gray}{$\pm$0.0}}} & 7.6\scriptsize{\textcolor{gray}{$\pm$0.8}} & 7.0\scriptsize{\textcolor{gray}{$\pm$0.8}} & 13.4\scriptsize{\textcolor{gray}{$\pm$1.1}} & 39.3\scriptsize{\textcolor{gray}{$\pm$1.5}} & \textbf{100.0\scriptsize{\textcolor{gray}{$\pm$0.0}}} \\ 
Cosine & \textbf{100.0\scriptsize{\textcolor{gray}{$\pm$0.0}}} & 0.8\scriptsize{\textcolor{gray}{$\pm$0.3}} & 1.5\scriptsize{\textcolor{gray}{$\pm$0.4}} & 6.7\scriptsize{\textcolor{gray}{$\pm$0.8}} & 32.8\scriptsize{\textcolor{gray}{$\pm$1.5}} & \textbf{100.0\scriptsize{\textcolor{gray}{$\pm$0.0}}} \\ 
\midrule
\textsc{Average} & \textbf{100.0\scriptsize{\textcolor{gray}{$\pm$0.0}}}& 14.4\scriptsize{\textcolor{gray}{$\pm$0.3}} & 32.0\scriptsize{\textcolor{gray}{$\pm$0.3}} & 38.6\scriptsize{\textcolor{gray}{$\pm$0.4}} & 55.2\scriptsize{\textcolor{gray}{$\pm$0.4}} & 99.8\scriptsize{\textcolor{gray}{$\pm$0.1}} \\

\bottomrule
\end{tabular}
\label{tab:arithmetic_accuracy_app}
\end{table}

\begin{table}[t]
\centering
\caption{Percent accuracy on reasoning tasks. Higher is Better. Bold indicates best performance.}
\scriptsize
\begin{tabular}{lccccccc}
\toprule
& OccamLlama & OccamLlama & Llama 3 & Llama 3 & GPT 3.5 & GPT 4o & GPT 4o \\ 
& 8B & 70B & 8b Instruct & 70B Instruct & Turbo &  & Code \\ 
\cmidrule(r){2-2}
\cmidrule(r){3-3}
\cmidrule(r){4-4}
\cmidrule(r){5-5}
\cmidrule(r){6-6}
\cmidrule(r){7-7}
\cmidrule(r){8-8}
AddSub & 91.6\scriptsize{\textcolor{gray}{$\pm$1.4}} & 96.5\scriptsize{\textcolor{gray}{$\pm$0.9}} & 93.4\scriptsize{\textcolor{gray}{$\pm$1.2}} & 97.2\scriptsize{\textcolor{gray}{$\pm$0.8}} & 95.4\scriptsize{\textcolor{gray}{$\pm$1.1}} & 97.0\scriptsize{\textcolor{gray}{$\pm$0.9}} & \textbf{97.5\scriptsize{\textcolor{gray}{$\pm$1.1}}} \\ 
GSM8K & 73.5\scriptsize{\textcolor{gray}{$\pm$1.2}} & 90.1\scriptsize{\textcolor{gray}{$\pm$0.8}} & 79.8\scriptsize{\textcolor{gray}{$\pm$1.1}} & 94.8\scriptsize{\textcolor{gray}{$\pm$0.6}} & 84.8\scriptsize{\textcolor{gray}{$\pm$1.0}} & \textbf{96.1\scriptsize{\textcolor{gray}{$\pm$0.5}}} & 94.0\scriptsize{\textcolor{gray}{$\pm$1.7}} \\ 
MultiArith & 99.2\scriptsize{\textcolor{gray}{$\pm$0.4}} & 98.5\scriptsize{\textcolor{gray}{$\pm$0.5}} & \textbf{99.8\scriptsize{\textcolor{gray}{$\pm$0.2}}} & 98.2\scriptsize{\textcolor{gray}{$\pm$0.5}} & 97.2\scriptsize{\textcolor{gray}{$\pm$0.7}} & 99.7\scriptsize{\textcolor{gray}{$\pm$0.2}} & 99.5\scriptsize{\textcolor{gray}{$\pm$0.5}} \\ 
MultiArith Float & \textbf{98.2\scriptsize{\textcolor{gray}{$\pm$0.5}}} & 97.7\scriptsize{\textcolor{gray}{$\pm$0.6}} & 57.3\scriptsize{\textcolor{gray}{$\pm$2.0}} & 76.3\scriptsize{\textcolor{gray}{$\pm$1.7}} & 77.3\scriptsize{\textcolor{gray}{$\pm$1.7}} & 96.2\scriptsize{\textcolor{gray}{$\pm$0.8}} & 89.5\scriptsize{\textcolor{gray}{$\pm$2.2}} \\ 
MATH401 & 85.0\scriptsize{\textcolor{gray}{$\pm$1.8}} & \textbf{89.5\scriptsize{\textcolor{gray}{$\pm$1.5}}} & 60.3\scriptsize{\textcolor{gray}{$\pm$2.4}} & 71.6\scriptsize{\textcolor{gray}{$\pm$2.3}} & 63.1\scriptsize{\textcolor{gray}{$\pm$2.4}} & 76.6\scriptsize{\textcolor{gray}{$\pm$2.1}} & 78.0\scriptsize{\textcolor{gray}{$\pm$2.9}} \\ 
Single Eq & 92.9\scriptsize{\textcolor{gray}{$\pm$1.1}} & 96.9\scriptsize{\textcolor{gray}{$\pm$0.8}} & 96.3\scriptsize{\textcolor{gray}{$\pm$0.8}} & 97.6\scriptsize{\textcolor{gray}{$\pm$0.7}} & 97.8\scriptsize{\textcolor{gray}{$\pm$0.6}} & 98.0\scriptsize{\textcolor{gray}{$\pm$0.6}} & \textbf{99.0\scriptsize{\textcolor{gray}{$\pm$0.7}}} \\ 
SVAMP & 88.6\scriptsize{\textcolor{gray}{$\pm$1.0}} & 93.2\scriptsize{\textcolor{gray}{$\pm$0.8}} & 86.3\scriptsize{\textcolor{gray}{$\pm$1.1}} & 94.5\scriptsize{\textcolor{gray}{$\pm$0.7}} & 87.8\scriptsize{\textcolor{gray}{$\pm$1.0}} & 96.2\scriptsize{\textcolor{gray}{$\pm$0.6}} & \textbf{96.5\scriptsize{\textcolor{gray}{$\pm$1.3}}} \\ 
\midrule
\textsc{Average} & 89.9\scriptsize{\textcolor{gray}{$\pm$1.2}} & \textbf{94.6\scriptsize{\textcolor{gray}{$\pm$0.9}}}& 81.9\scriptsize{\textcolor{gray}{$\pm$1.5}} & 90.0\scriptsize{\textcolor{gray}{$\pm$1.2}} & 86.2\scriptsize{\textcolor{gray}{$\pm$1.4}} & 94.2\scriptsize{\textcolor{gray}{$\pm$1.0}} & 93.4\scriptsize{\textcolor{gray}{$\pm$1.7}} \\

\bottomrule
\end{tabular}
\label{tab:reasoning_accuracy}
\end{table}

\begin{table}[t]
\centering
\caption{Relative error (\%) on arithmetic tasks. Lower is Better. Bold indicates best performance.}
\scriptsize
\begin{tabular}{lcccccc}
\toprule
& OccamLlama & Llama 2 & Llama 3 & GPT 3.5 & GPT 4o & GPT 4o \\ 
&  & 7B Chat & 8b Instruct & Turbo &  & Code \\ 
\cmidrule(r){2-2}
\cmidrule(r){3-3}
\cmidrule(r){4-4}
\cmidrule(r){5-5}
\cmidrule(r){6-6}
\cmidrule(r){7-7}
Addition (3) & \textbf{0.0\scriptsize{\textcolor{gray}{$\pm$0.0}}} & 50.5\scriptsize{\textcolor{gray}{$\pm$10.9}} & 3.2\scriptsize{\textcolor{gray}{$\pm$1.9}} & 0.3\scriptsize{\textcolor{gray}{$\pm$0.1}} & \textbf{0.0\scriptsize{\textcolor{gray}{$\pm$0.0}}} &  \\ 
Addition (5) & \textbf{0.0\scriptsize{\textcolor{gray}{$\pm$0.0}}} & 113.0\scriptsize{\textcolor{gray}{$\pm$21.1}} & 23.7\scriptsize{\textcolor{gray}{$\pm$4.0}} & 4.6\scriptsize{\textcolor{gray}{$\pm$1.8}} & 0.0\scriptsize{\textcolor{gray}{$\pm$0.0}} &  \\ 
Addition (7) & \textbf{0.0\scriptsize{\textcolor{gray}{$\pm$0.0}}} & 310.3\scriptsize{\textcolor{gray}{$\pm$97.0}} & 78.1\scriptsize{\textcolor{gray}{$\pm$16.2}} & 4.0\scriptsize{\textcolor{gray}{$\pm$1.4}} & 1.0\scriptsize{\textcolor{gray}{$\pm$0.9}} & \textbf{0.0\scriptsize{\textcolor{gray}{$\pm$0.0}}} \\ 
\\
Subtraction (3) & \textbf{0.0\scriptsize{\textcolor{gray}{$\pm$0.0}}} & 66.2\scriptsize{\textcolor{gray}{$\pm$18.7}} & 4.1\scriptsize{\textcolor{gray}{$\pm$0.8}} & 3.8\scriptsize{\textcolor{gray}{$\pm$0.7}} & 0.4\scriptsize{\textcolor{gray}{$\pm$0.1}} &  \\ 
Subtraction (5) & \textbf{0.0\scriptsize{\textcolor{gray}{$\pm$0.0}}} & 173.6\scriptsize{\textcolor{gray}{$\pm$67.1}} & 29.4\scriptsize{\textcolor{gray}{$\pm$4.3}} & 38.3\scriptsize{\textcolor{gray}{$\pm$16.5}} & 3.5\scriptsize{\textcolor{gray}{$\pm$0.6}} &  \\ 
Subtraction (7) & \textbf{0.0\scriptsize{\textcolor{gray}{$\pm$0.0}}} & 222.3\scriptsize{\textcolor{gray}{$\pm$54.4}} & 65.6\scriptsize{\textcolor{gray}{$\pm$12.9}} & 44.6\scriptsize{\textcolor{gray}{$\pm$31.3}} & 5.4\scriptsize{\textcolor{gray}{$\pm$0.7}} & 0.3\scriptsize{\textcolor{gray}{$\pm$0.3}} \\ 
\\
Multiplication (3) & \textbf{0.0\scriptsize{\textcolor{gray}{$\pm$0.0}}} & 7.6\scriptsize{\textcolor{gray}{$\pm$0.7}} & 1.9\scriptsize{\textcolor{gray}{$\pm$0.5}} & 1.8\scriptsize{\textcolor{gray}{$\pm$0.4}} & 0.1\scriptsize{\textcolor{gray}{$\pm$0.1}} &  \\ 
Multiplication (5) & \textbf{0.0\scriptsize{\textcolor{gray}{$\pm$0.0}}} & 84.9\scriptsize{\textcolor{gray}{$\pm$0.9}} & 46.7\scriptsize{\textcolor{gray}{$\pm$1.7}} & 19.2\scriptsize{\textcolor{gray}{$\pm$3.7}} & 1.8\scriptsize{\textcolor{gray}{$\pm$0.4}} &  \\ 
Multiplication (7) & \textbf{0.0\scriptsize{\textcolor{gray}{$\pm$0.0}}} & 98.9\scriptsize{\textcolor{gray}{$\pm$0.2}} & 74.9\scriptsize{\textcolor{gray}{$\pm$1.8}} & 90.1\scriptsize{\textcolor{gray}{$\pm$24.2}} & 4.4\scriptsize{\textcolor{gray}{$\pm$0.6}} & 1.0\scriptsize{\textcolor{gray}{$\pm$0.7}} \\ 
\\
Division (3) & 0.1\scriptsize{\textcolor{gray}{$\pm$0.0}} & 1346.2\scriptsize{\textcolor{gray}{$\pm$275.4}} & 1.3\scriptsize{\textcolor{gray}{$\pm$0.9}} & 1.1\scriptsize{\textcolor{gray}{$\pm$0.3}} & \textbf{0.0\scriptsize{\textcolor{gray}{$\pm$0.0}}} &  \\ 
Division (5) & 0.2\scriptsize{\textcolor{gray}{$\pm$0.1}} & 174156.6\scriptsize{\textcolor{gray}{$\pm$31687.6}} & 9.5\scriptsize{\textcolor{gray}{$\pm$1.8}} & 0.7\scriptsize{\textcolor{gray}{$\pm$0.2}} & \textbf{0.1\scriptsize{\textcolor{gray}{$\pm$0.0}}} &  \\ 
Division (7) & 0.1\scriptsize{\textcolor{gray}{$\pm$0.0}} & 22032920.9\scriptsize{\textcolor{gray}{$\pm$3642549.7}} & 225.3\scriptsize{\textcolor{gray}{$\pm$142.1}} & 0.3\scriptsize{\textcolor{gray}{$\pm$0.1}} & 0.0\scriptsize{\textcolor{gray}{$\pm$0.0}} & \textbf{0.0\scriptsize{\textcolor{gray}{$\pm$0.0}}} \\ 
\\
Square Root (3) & \textbf{0.0\scriptsize{\textcolor{gray}{$\pm$0.0}}} & 8.9\scriptsize{\textcolor{gray}{$\pm$1.1}} & 1.1\scriptsize{\textcolor{gray}{$\pm$0.3}} & 0.2\scriptsize{\textcolor{gray}{$\pm$0.0}} & 0.0\scriptsize{\textcolor{gray}{$\pm$0.0}} &  \\ 
Square Root (5) & \textbf{0.0\scriptsize{\textcolor{gray}{$\pm$0.0}}} & 72.8\scriptsize{\textcolor{gray}{$\pm$4.8}} & 12.6\scriptsize{\textcolor{gray}{$\pm$1.7}} & 0.1\scriptsize{\textcolor{gray}{$\pm$0.0}} & 0.0\scriptsize{\textcolor{gray}{$\pm$0.0}} &  \\ 
Square Root (7) & 0.0\scriptsize{\textcolor{gray}{$\pm$0.0}} & 207.8\scriptsize{\textcolor{gray}{$\pm$21.8}} & 15.4\scriptsize{\textcolor{gray}{$\pm$1.4}} & 8.8\scriptsize{\textcolor{gray}{$\pm$0.9}} & 4.8\scriptsize{\textcolor{gray}{$\pm$2.0}} & \textbf{0.0\scriptsize{\textcolor{gray}{$\pm$0.0}}} \\ 
\\
Exponential & 0.3\scriptsize{\textcolor{gray}{$\pm$0.0}} & 422.6\scriptsize{\textcolor{gray}{$\pm$82.6}} & 11.7\scriptsize{\textcolor{gray}{$\pm$0.7}} & 2.3\scriptsize{\textcolor{gray}{$\pm$0.9}} & 0.1\scriptsize{\textcolor{gray}{$\pm$0.0}} & \textbf{0.0\scriptsize{\textcolor{gray}{$\pm$0.0}}} \\ 
Logarithm & 0.0\scriptsize{\textcolor{gray}{$\pm$0.0}} & 138.3\scriptsize{\textcolor{gray}{$\pm$11.9}} & 40.2\scriptsize{\textcolor{gray}{$\pm$1.3}} & 6.4\scriptsize{\textcolor{gray}{$\pm$4.0}} & 0.1\scriptsize{\textcolor{gray}{$\pm$0.0}} & \textbf{0.0\scriptsize{\textcolor{gray}{$\pm$0.0}}} \\ 
Sine & 0.1\scriptsize{\textcolor{gray}{$\pm$0.0}} & 515.5\scriptsize{\textcolor{gray}{$\pm$68.1}} & 279.2\scriptsize{\textcolor{gray}{$\pm$21.9}} & 125.4\scriptsize{\textcolor{gray}{$\pm$12.3}} & 3.2\scriptsize{\textcolor{gray}{$\pm$0.7}} & \textbf{0.1\scriptsize{\textcolor{gray}{$\pm$0.0}}} \\ 
Cosine & 0.1\scriptsize{\textcolor{gray}{$\pm$0.0}} & 581.9\scriptsize{\textcolor{gray}{$\pm$199.7}} & 346.1\scriptsize{\textcolor{gray}{$\pm$80.4}} & 188.5\scriptsize{\textcolor{gray}{$\pm$22.5}} & 2.8\scriptsize{\textcolor{gray}{$\pm$0.5}} & \textbf{0.0\scriptsize{\textcolor{gray}{$\pm$0.0}}} \\ 
\midrule
\textsc{Average} & \textbf{0.1\scriptsize{\textcolor{gray}{$\pm$0.0}}}& 1169210.8\scriptsize{\textcolor{gray}{$\pm$194953.4}} & 66.9\scriptsize{\textcolor{gray}{$\pm$8.8}} & 28.4\scriptsize{\textcolor{gray}{$\pm$2.7}} & 1.5\scriptsize{\textcolor{gray}{$\pm$0.1}} & 0.2\scriptsize{\textcolor{gray}{$\pm$0.1}} \\

\bottomrule
\end{tabular}
\label{tab:arithmetic_error_app}
\end{table}

\section{Example OccamLlama Generations}
In this section, we include example OccamLlama 8B generations from the MATH401 and MultiArith-Float datasets. We randomly selected three examples for each dataset. OccamNet outputs are included in \textcolor{green}{green}. We omit prompt formatting to save space. Similarly, although outputs from OccamNet are always followed by ``\textbackslash n\textbackslash n,'' we omit these newlines to save space, instead adding a period and space after each OccamNet generation.

By chance, all six responses happen to be correct.

We also include three miscellaneous prompts and OccamLlama 8B generations to demonstrate OccamLlama's ability to handle more challenging prompts.

\subsection{MATH401}
\textbf{Example 1}

Prompt:
\begin{quotation}
    \textit{\textcolor{gray}{7\textasciicircum 4=}}
\end{quotation}

Response:
\begin{quotation}
    \textit{\textcolor{gray}{\textcolor{green}{2401}. 7 \textasciicircum 4 = 7 × 7 × 7 × 7 = 2401}}
\end{quotation}

\textbf{Example 2}

Prompt:
\begin{quotation}
    \textit{\textcolor{gray}{6**3=}}
\end{quotation}
Response:
\begin{quotation}
    \textit{\textcolor{gray}{\textcolor{green}{216}. 6 \textasciicircum 3 = 6 × 6 × 6 = 216}}
\end{quotation}

\textbf{Example 3}

Prompt:
\begin{quotation}
    \textit{\textcolor{gray}{63.0+0.074=}}
\end{quotation}
Response:
\begin{quotation}
    \textit{\textcolor{gray}{63.074. 63.0 + 0.074 = \textcolor{green}{63.074}. So the answer is 63.074.}}
\end{quotation}

\subsection{MultiArith-Float}
\textbf{Example 1}

Prompt:
\begin{quotation}
    \textit{\textcolor{gray}{Mike had 30 video games but 6 of them weren't working. If he wanted to sell the working games for 85.53 each, how much money could he earn?}}
\end{quotation}
Response:
\begin{quotation}
    \textit{\textcolor{gray}{Mike had 30 video games. 6 weren't working, so he had 30 - 6 = \textcolor{green}{24}. He can sell 24 games for 85.53 each. 24 x 85.53 is \textcolor{green}{2052.720}. So Mike could earn 2052.72 dollars.}}
\end{quotation}

\textbf{Example 2}

Prompt:
\begin{quotation}
    \textit{\textcolor{gray}{While playing a trivia game, Adam answered 6 questions correct in the first half and 3 questions correct in the second half. If each question was worth 59.08 points, what was his final score?}}
\end{quotation}
Response:
\begin{quotation}
    \textit{\textcolor{gray}{Adam got 6 questions correct in the first half, and 3 questions correct in the second half. So he got 6 + 3 = \textcolor{green}{9}. Each question was worth 59.08 points. So his score is 9 * 59.08 = \textcolor{green}{531.720}. The answer is 531.72.}}
\end{quotation}

\textbf{Example 3}

Prompt:
\begin{quotation}
    \textit{\textcolor{gray}{Lana's favorite band was holding a concert where tickets were 181.92 dollars each. Lana bought 5 tickets for herself and her friends and 4 extra tickets in case anyone else wanted to go. How much did she spend?}}
\end{quotation}
Response:
\begin{quotation}
    \textit{\textcolor{gray}{Lana bought 5 tickets for herself and her friends. Then 4 extra tickets. So in total she bought 5 + 4 = \textcolor{green}{9}. Each ticket was 181.92 dollars. So 9 tickets will be 9 x 181.92 = \textcolor{green}{1637.280}. The answer is 1637.280.}}
\end{quotation}

\subsection{Miscellaneous prompts}
These miscellaneous prompts demonstrate OccamLlama's ability to handle more challenging prompts. Examples 1 and 2 show that OccamLlama can handle prompts where the numbers are provided as words and not as decimals. This works because we can choose to parse strings such as ``three'' and ``four'' as 3 and 4, respectively, when parsing inputs to OccamNet. Example 3 shows that OccamLlama can handle prompts in which many irrelevant numbers are included. This works because OccamLlama uses Llama to reason and select the correct numbers to use before feeding these numbers into OccamNet.

\textbf{Example 1}

Prompt:
\begin{quotation}
    \textit{\textcolor{gray}{Six minus seven =?}}
\end{quotation}
Response:
\begin{quotation}
    \textit{\textcolor{gray}{\textcolor{green}{-1}. Six minus seven is equal to -1.}}
\end{quotation}

\textbf{Example 2}

Prompt:
\begin{quotation}
    \textit{\textcolor{gray}{Three + four = ?}}
\end{quotation}
Response:
\begin{quotation}
    \textit{\textcolor{gray}{\textcolor{green}{7}. 3 + 4 = \textcolor{green}{7}. Let me know if you have any other questions!}}
\end{quotation}

\textbf{Example 3}

Prompt:
\begin{quotation}
    \textit{\textcolor{gray}{In 1932, 100 planets were discovered and in 1933, another 200 were found. What is the total number of planets discovered between 1932 and 1933?}}
\end{quotation}
Response:
\begin{quotation}
    \textit{\textcolor{gray}{The number of planets discovered in 1932 was 100. The number of planets discovered in 1933 was 200. So the total number of planets discovered is 100 + 200 = \textcolor{green}{300}. 300.}}
\end{quotation}

\section{OccamLLM Robustness Experiments} \label{app: robustness}
We find that OccamLlama displays remarkable generalization capabilities on out-of-distribution problems. To demonstrate this, we show below two out-of-distribution tasks on which OccamLlama performs remarkably well.

\subsection{Non-textual Training}
First, we train the OccamNet decoder \textit{from scratch}, using \textit{only numeric expressions} and \textit{absolutely no text at all}. This means that any problem with text, such as a word problem, is far out-of-distribution of the OccamNet decoder's training data. We test this model (using the standard router), which we denote OccamLlama 8B Arith, on the mathematical reasoning benchmarks and obtain remarkably good results, shown in Table \ref{tab:reasoning_arith}.

\begin{table}[t]
\centering
\caption{Accuracy on reasoning tasks. Higher is Better.}
\footnotesize
\begin{tabular}{lccccccc}
\toprule
& OccamLlama & OccamLlama & Llama 2 & Llama 3 & GPT 3.5 & GPT 4o & GPT 4o \\ 
& 8B & 8B Arith & 7B Chat & 8b Instruct & Turbo &  & Code \\ 
\cmidrule(r){2-2}
\cmidrule(r){3-3}
\cmidrule(r){4-4}
\cmidrule(r){5-5}
\cmidrule(r){6-6}
\cmidrule(r){7-7}
\cmidrule(r){8-8}
AddSub & 91.6\scriptsize{\textcolor{gray}{$\pm$1.4}} & 92.7\scriptsize{\textcolor{gray}{$\pm$1.3}} & 78.0\scriptsize{\textcolor{gray}{$\pm$2.1}} & 93.4\scriptsize{\textcolor{gray}{$\pm$1.2}} & 95.4\scriptsize{\textcolor{gray}{$\pm$1.1}} & 97.0\scriptsize{\textcolor{gray}{$\pm$0.9}} & \textbf{97.5\scriptsize{\textcolor{gray}{$\pm$1.1}}} \\ 
GSM8K & 73.5\scriptsize{\textcolor{gray}{$\pm$1.2}} & 71.6\scriptsize{\textcolor{gray}{$\pm$1.2}} & 36.0\scriptsize{\textcolor{gray}{$\pm$1.3}} & 79.8\scriptsize{\textcolor{gray}{$\pm$1.1}} & 84.8\scriptsize{\textcolor{gray}{$\pm$1.0}} & \textbf{96.1\scriptsize{\textcolor{gray}{$\pm$0.5}}} & 94.0\scriptsize{\textcolor{gray}{$\pm$1.7}} \\ 
MultiArith & 99.2\scriptsize{\textcolor{gray}{$\pm$0.4}} & 98.5\scriptsize{\textcolor{gray}{$\pm$0.5}} & 76.0\scriptsize{\textcolor{gray}{$\pm$1.7}} & \textbf{99.8\scriptsize{\textcolor{gray}{$\pm$0.2}}} & 97.2\scriptsize{\textcolor{gray}{$\pm$0.7}} & 99.7\scriptsize{\textcolor{gray}{$\pm$0.2}} & 99.5\scriptsize{\textcolor{gray}{$\pm$0.5}} \\ 
MultiArith Float & \textbf{98.2\scriptsize{\textcolor{gray}{$\pm$0.5}}} & 95.3\scriptsize{\textcolor{gray}{$\pm$0.9}} & 23.3\scriptsize{\textcolor{gray}{$\pm$1.7}} & 57.3\scriptsize{\textcolor{gray}{$\pm$2.0}} & 77.3\scriptsize{\textcolor{gray}{$\pm$1.7}} & 96.2\scriptsize{\textcolor{gray}{$\pm$0.8}} & 89.5\scriptsize{\textcolor{gray}{$\pm$2.2}} \\ 
MATH401 & 85.0\scriptsize{\textcolor{gray}{$\pm$1.8}} & \textbf{85.8\scriptsize{\textcolor{gray}{$\pm$1.7}}} & 43.9\scriptsize{\textcolor{gray}{$\pm$2.5}} & 60.3\scriptsize{\textcolor{gray}{$\pm$2.4}} & 63.1\scriptsize{\textcolor{gray}{$\pm$2.4}} & 76.6\scriptsize{\textcolor{gray}{$\pm$2.1}} & 78.0\scriptsize{\textcolor{gray}{$\pm$2.9}} \\ 
Single Eq & 92.9\scriptsize{\textcolor{gray}{$\pm$1.1}} & 92.1\scriptsize{\textcolor{gray}{$\pm$1.2}} & 79.1\scriptsize{\textcolor{gray}{$\pm$1.8}} & 96.3\scriptsize{\textcolor{gray}{$\pm$0.8}} & 97.8\scriptsize{\textcolor{gray}{$\pm$0.6}} & 98.0\scriptsize{\textcolor{gray}{$\pm$0.6}} & \textbf{99.0\scriptsize{\textcolor{gray}{$\pm$0.7}}} \\ 
SVAMP & 88.6\scriptsize{\textcolor{gray}{$\pm$1.0}} & 88.8\scriptsize{\textcolor{gray}{$\pm$1.0}} & 61.5\scriptsize{\textcolor{gray}{$\pm$1.5}} & 86.3\scriptsize{\textcolor{gray}{$\pm$1.1}} & 87.8\scriptsize{\textcolor{gray}{$\pm$1.0}} & 96.2\scriptsize{\textcolor{gray}{$\pm$0.6}} & \textbf{96.5\scriptsize{\textcolor{gray}{$\pm$1.3}}} \\ 
\midrule
\textsc{Average} & 89.9\scriptsize{\textcolor{gray}{$\pm$1.1}} & 89.3\scriptsize{\textcolor{gray}{$\pm$1.1}} & 56.8\scriptsize{\textcolor{gray}{$\pm$1.8}} & 81.9\scriptsize{\textcolor{gray}{$\pm$1.3}} & 86.2\scriptsize{\textcolor{gray}{$\pm$1.2}} & \textbf{94.2\scriptsize{\textcolor{gray}{$\pm$0.8}}}& 93.4\scriptsize{\textcolor{gray}{$\pm$1.5}} \\

\bottomrule
\end{tabular}
\label{tab:reasoning_arith}
\end{table}

The OccamLlama 8B Arith performs on par with the model trained with both numbers and text, even achieving higher accuracy on some benchmarks. This shows that the OccamLLM framework is robust, and points towards the fact that the representations of arithmetic that are built in the transformer body of the LLM and extracted by the OccamLLM Decoder are very general.

In contrast, we expect that finetuning Llama to perform arithmetic using only numeric examples and no text whatsoever would lead to extreme catastrophic forgetting and poor arithmetic performance on word problems. As such, we believe this data shows a remarkable generalization and robustness of OccamLLM.

\subsection{Multilingual Reasoning}

\begin{figure}
    \centering
    \includegraphics[width=\textwidth]{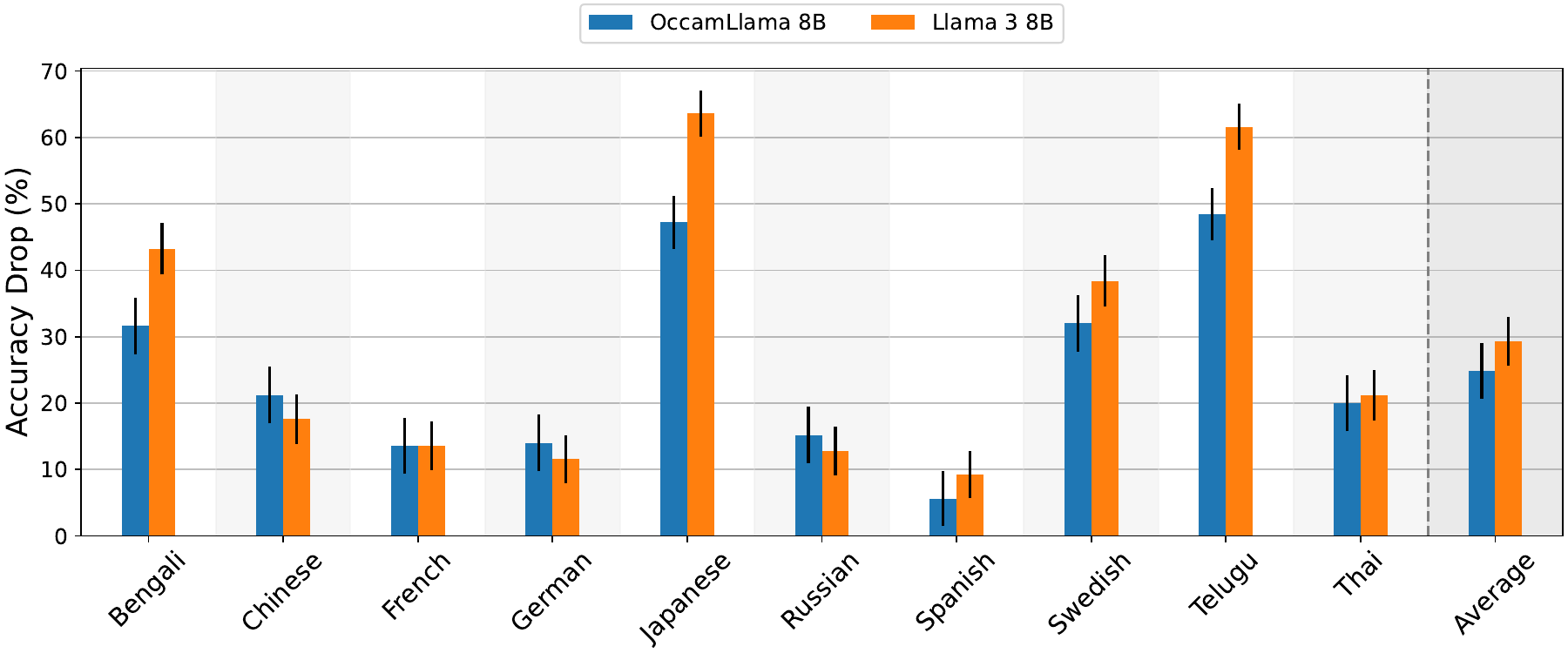}
    \caption{Model performance degradation for each language relative to English in the MGSM dataset. OccamLlama 8B's performance degradation is considerably less than Llama 3 8B's performance degradation, demonstrating strong multilingual and generalization capabilities.}
    \label{fig: mgsm}
\end{figure}

To further demonstrate OccamLlama's  generalization capabilities and also show that OccamLlama can handle non-English generation, we tested OccamLlama on the Multilingual Grade School Math Benchmark (MGSM) \cite{shi2022languagemodelsmultilingualchainofthought}, a dataset consisting of GSM8K translated into 10 languages (Bengali, Chinese, French, German, Japanese, Russian, Spanish, Swahili, Telugu, and Thai). For these experiments, we prompted the LLMs to write their answers in the same language as the problem statement. Otherwise, the LLM would typically respond always in English, defeating the purpose of the experiment. We compute the drop in accuracy when switching from English to another language, given by the accuracy of a model on the English dataset minus the accuracy of the model on the dataset in a given language. The results are shown in Figure \ref{fig: mgsm}.

The table above shows that OccamLlama and Llama have similar performance drops between the English dataset and each non-English language dataset. On most languages and on average, OccamLlama has a smaller performance drop than Llama. The fact that OccamLlama (the decoders for which have never been trained on other languages) has a smaller average out-of-distribution performance drop than Llama (a model trained on over 750 billion tokens of non-English text) is in our opinion quite remarkable.

We believe that this test demonstrates OccamLlama's ability to handle many languages and its robustness against out-of-distribution data.

\section{Alternative Architectures and Losses}\label{app: alternative architectures}

\subsection{Alternative Architectures}
As discussed in the main text, although OccamLLM works most naturally with OccamNet, it can also work with other symbolic architectures such as the EQL network \cite{EQLOriginal, EQLWithDivision}, or architectures that can represent probability distributions over symbolic expressions, such as transformers \cite{attentionAllYouNeed} or recurrent neural networks (RNNs) \cite{rnns}.

However, in practice we believe OccamNet is the most effective architecture for this use case. We find that because EQL does not represent a probability distribution over functions, it easily gets stuck in local minima. 

Regarding transformers and RNNs, we believe that OccamNet possesses a key advantage of being interpretable; simply by looking at the weights, it is possible for a human to determine which functions OccamNet assigns a high probability. We believe that this interpretability will make OccamNet easy for a decoder to initialize with the desired distribution. On the other hand, an RNN or transformer have substantially more complex relations between the weights and corresponding probability distribution, which we hypothesize would make learning a decoder for such models difficult.

This leads us to a key point: transformers and RNNs are effective for modeling complex multimodal distributions, but for this problem, we want to select a single function for each token, so the extra expressivity of these models is unneeded and likely detrimental to performance. We believe that OccamNet, a much simpler architecture, enables better parameter efficiency and performance.

\subsection{Alternative Losses}\label{app: alt losses}
In this section we discuss alternative possible losses and how we arrived at the loss in Equation \ref{eq: Loss}.

We considered two loss functions which are natural when optimizing a probability distribution: 1) a cross-entropy loss, and 2) a REINFORCE \cite{reinforce} loss. Each of these requires only a slight modification to reach Equation \ref{eq: Loss}. This discussion thus illustrates how our loss combines benefits from both the cross-entropy and the reinforcement-learning losses.

\paragraph{Cross-Entropy Loss}
The cross-entropy loss is effective at modeling probability distributions. Given a ground truth distribution $q_x[f]$ conditioned on the input text $x$, the cross-entropy loss is given by \begin{equation}
    \mathcal{L}(x,y;W) = - \sum_f q_x[f]\log p_W[f].
\end{equation} Unfortunately, for OccamLLM, the ground-truth distribution $q_x[f]$ is not uniquely specified. In particular the only constraints on $q_x[f]$ are that it is is normalized and satisfies $q_x[f] = 0$ if $f$ is not the desired function (i.e., $f(x)\neq y$). Since the same function can be represented in many ways in the OccamNet network (a property true of many function representations), multiple $f$ may satisfy $f(x) = y$, so $q_x$ is underparametrized.

The most natural choice for $q_x$ is to weight each valid function equally:
\begin{equation}
    q_x[f] = \begin{cases}
        c_x & \text{if } f(x) = y\\
        0 & \text{otherwise}
    \end{cases}
\end{equation}
where $c_x$ is a constant chosen such that $q_x$ is normalized, given by the inverse of the number of functions $f$ satisfying $f(x) = y$. However, determining $c_x$ requires testing every possible function $f$, which may be infeasible for large OccamNet networks. Further, this $q_x$ requires OccamNet to learn a superposition of functions, which may be challenging given its relatively low parameter count.

Another option is to choose a canonical form $f^*$ for each function and to set $q_x$ to be a 1-hot distribution that is nonzero only at $f^*.$ Although this removes the challenge of learning a superposition, it still requires sampling nearly all functions in OccamNet due to the sparsity of $q_x.$

Ideally, we would like to find a $q_x$ with the following conditions:
\begin{itemize}
    \item It enables the cross-entropy loss to be calculated by sampling from OccamNet. This allows us to avoid needing to iterate through and evaluate every $f(x)$ each time we compute the loss, since we can instead obtain a Monte-Carlo estimate.
    
    \item It is minimized when $p_W$ is a 1-hot probability distribution. This ensures that OccamNet can represent the optimal distribution.

    \item It has $q_x[f] \neq 0$ for all $f$ satisfying $f(x)=y$. This improves sample-efficiency by increasing the probability of sampling an $f$ with $q_x[f] > 0$.
\end{itemize}

A solution is to set 
\begin{equation}
    q_x[f] = \begin{cases}
        c_x p_W[f] & \text{if } f(x) = y\\
        0 & \text{otherwise}
    \end{cases}
\end{equation}
where $c_x$ is chosen such that $q_x$ is normalized. This gives a loss
\begin{align*}
    \mathcal{L}(x,y;W) &= - \sum_f q_x[f]\cdot \log p_W[f]\\
    &= - \sum_f c_x p_W[f] \cdot \delta(f(x)-y) \cdot \log p_W[f]\\
    &\approx - \frac{c_x}{N} \sum_{f\sim p_W} \delta(f(x)-y) \cdot \log p_W[f]\\
    &\approx - \frac{\sum_{f\sim p_W} \delta(f(x)-y) \cdot \log p_W[f]}{\sum_{f\sim p_W} \delta(f(x)-y)},
\end{align*} where \begin{equation}
    \delta(f(x)-y) = \begin{cases}
        1 & \text{if } f(x) = y\\
        0 & \text{otherwise}
    \end{cases}
\end{equation} and in the last step we used the fact that $c_x$ can be approximated as
\begin{equation*}
    c_x  = \frac{1}{\sum_f p_W[f] \delta(f(x)-y)} \approx \frac{N}{\sum_{f\sim p_W} \delta(f(x)-y)}.
\end{equation*} This loss is easily computed by sampling from $p_W$, it satisfies $q_W>0$ for all $f$ satisfying $f(x) = y,$ and it is minimized when $p_W$ is a delta function centered at any $f$ satisfying $f(x) = y,$ as desired.

Note that 
\begin{equation}
    \mathcal{L}(x,y;W) = - \frac{\sum_{f\sim p_W} \delta(f(x)-y) \cdot \log p_W[f]}{\sum_{f\sim p_W} \delta(f(x)-y)}
\end{equation}
is exactly the loss given in Equation \ref{eq: Loss} with $R(f(x),y) = \delta(f(x)-y)$. Thus, we have shown how Equation \ref{eq: Loss} can be interpreted as a cross-entropy loss. Equation \ref{eq: Loss} with general $R(f(x),y)$ can be seen as a cross-entropy loss with a ``smoothed'' ground truth distribution $q_x$ given by $q_x \propto p_W[f] \cdot R(f(x),y).$

\paragraph{REINFORCE Loss}
Reinforcement-learning losses are effective for exploring large search spaces. We use a modification of the REINFORCE \cite{reinforce} loss because it is relatively simple to implement. Future work could explore more sophisticated variants of this algorithm, such as Proximal Policy Optimization \cite{ppo}.

The standard REINFORCE loss applied to OccamLLM gives 
\begin{equation*}
    \mathcal{L}(x,y;W) = - \frac{1}{N} \sum_{f\sim p_W} R(f(x),y) \cdot \log p_W[f].
\end{equation*}
Note that for sparse $R$, there will be very few nonzero $R(f(x),y)$ sampled, so, since we are dividing by $N$, the gradient signal will be small. We modify REINFORCE by dividing by the the sum of the rewards for all samples instead of by $N$ to ensure that correct functions sampled only a few times still receive a large training step update. This once again produces Equation \ref{eq: Loss}.

We find that using a delta function for our reward is most effective because it most accurately represents the sparse reward of the problem. Further, as shown above, this loss provides a Monte-Carlo estimate of the the cross entropy loss. Due to the sparse reward, many samples may initially be required to obtain an accurate estimate of the loss. However, as OccamNet approaches the desired distribution, the loss's sample efficiency will improve.

\section{Background on OccamNet}\label{app: background on occamnet}

This section is heavily modified from \cite{OccamNet}.

We divide this section into the following subsections:
\begin{enumerate}
    \item In Section \ref{sec:occamnet_architecture}, we describe OccamNet's architecture in more detail.
    \item In Section \ref{sec:sampling}, we describe OccamNet's sampling process.
    \item In Section \ref{sec:probability}, we describe OccamNet's probability distribution.
    \item In Section \ref{sec:initialization}, we describe OccamNet's initialization process.
\end{enumerate}

\subsection{OccamNet Architecture}
\label{sec:occamnet_architecture}

\begin{figure*}
    \centering
    
    \includegraphics[width=\linewidth]{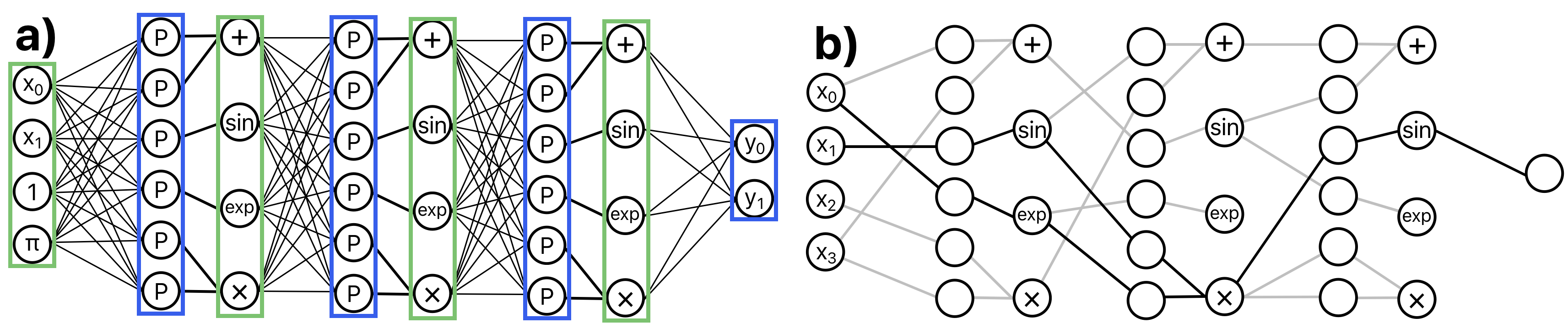}
    
    \caption{$a)$ An example OccamNet, with image layers boxed in green and arguments layers boxed in blue. We denote the inputs as the 0th image layer and the outputs as the $(L+1)$th arguments layer.  Nodes in the arguments layers are represented with a P because of their probabilistic nature. $b)$ A demonstration of the dropped connections from sampled paths in OccamNet. All light grey paths are dropped from the final symbolic form of the sampled function because they are not directly connected to the outputs.}
    
    \label{fig:DroppedConnections}
\end{figure*}

As described in the main text, we start from a predefined collection of $N$ primitive functions $\mathcal{P}$. OccamNet represents a distribution over compositions of functions in $\mathcal{P}.$ From now on, we denote the $i$th primitive function in the $l$th layer as $\phi_i^{(l)}$. We begin indexing the primitives from 0 and the layers from 1, because we treat the inputs as the 0th layer. So, for example, in Figure \ref{fig:DroppedConnections}a, $\phi_2^{(1)} = \phi_2^{(2)} = \phi_2^{(3)} = \mathsf{sin}$.

Each OccamNet layer consists of two sublayers, which we denote the \textit{arguments} and \textit{image} sublayers, shown in Figure \ref{fig:DroppedConnections}a. For an $L$-layer OccamNet, each of these sublayers is reproduced $L$ times. The $l$th softmax layer connects the $(l-1)$th image layer with the $l$th arguments layer. For $1\le l\le L$, we denote the $l$th arguments sublayer hidden state as $\widetilde{\mathbf{h}}^{(l)}$ and the $l$th image sublayer hidden state as $\mathbf{h}^{(l)}$. So, $\widetilde{\mathbf{h}}^{(2)}$ would represent the middle layer of nodes labeled $P$ in Figure \ref{fig:DroppedConnections}a. We further write
\begin{gather}
\mathbf{\widetilde h}^{(l)}=\left[\widetilde h^{(l)}_1,\dots, \widetilde h^{(l)}_{M^{(l)}}\right]^\top, \;\;
\mathbf{h}^{(l)}=\left[h^{(l)}_1,\dots,h^{(l)}_{N^{(l)}}\right]^\top,
\end{gather} where \begin{equation*}
    M^{(l)} = \sum_{0\leq k < N^{(l)}} \alpha\left[\phi_k^{(l)}\right],
\end{equation*} $N^{(l)}$ is the number of primitives in layer $l$, and $\alpha[\phi]$ is the arity of function $\phi.$ We also define $\mathbf{h}^{(0)}$ to be the input layer (an image sublayer) and $\mathbf{\widetilde h}^{(L+1)}$ to be the output layer (an arguments sublayer).

\begin{figure}[t]
    \centering
    \includegraphics[width=\textwidth]{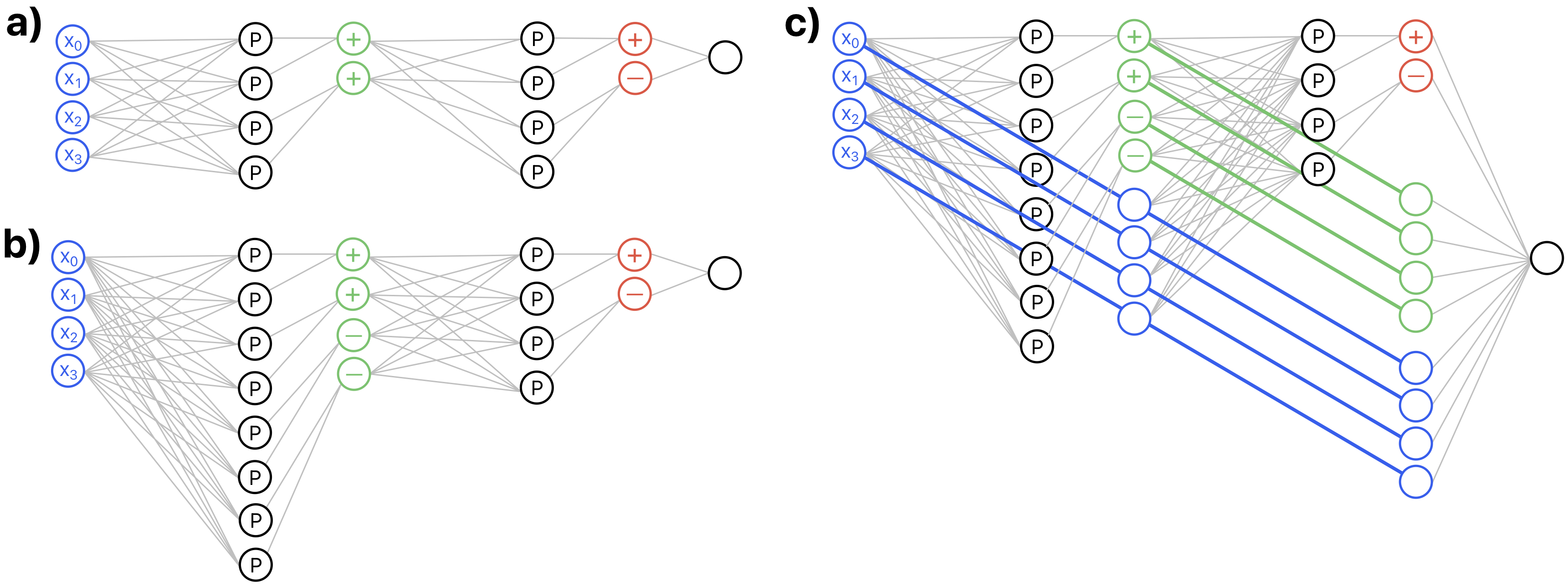}
    \caption{The progression of enhancements leading to a Complete OccamNet from a standard OccamNet. a) A standard OccamNet without repeated activations or skip connections. b) The same OccamNet as in a) with activations repeated in earlier layers. c) The same OccamNet as in b) with added skip connections. This is a Complete OccamNet.}
    \label{fig:mesh2}
\end{figure}

In a standard OccamNet layer, each primitive is repeated exactly once in each layer. However, in Complete OccamNet, each primitive in the $l$th layer is repeated $A^{L-l}$ times, where $A$ is the maximum arity of the primitives. This is shown in Figure \ref{fig:mesh2} in the transition from \ref{fig:mesh2}a to \ref{fig:mesh2}b. Complete OccamNet also concatenates each image layer to the next image layer, as shown in Figure \ref{fig:mesh2}c.

\subsection{Sampling from OccamNet}
\label{sec:sampling}

In this section, we more carefully describe OccamNet's sampling process. We sample a connection to each arguments layer node from the distribution given by the softmax of the softmax-layer weights leading to that node. In particular, if $\mathbf{w}_i^{(l)}$ are the weights of the $l$th softmax layer leading to the $i$th node of the $l$th argument's layer, when we sample we produce a sparse matrix
\begin{equation}
\mathsf{SAMPLE}\left(\begin{bmatrix} {\mathsf{softmax}(\mathbf{w}_1^{(l)})} \\ \vdots \\ {\mathsf{softmax}(\mathbf{w}_{M^{(l)}}^{(l)})}\end{bmatrix}\right)
\end{equation} where the $\mathsf{SAMPLE}$ function samples a one-hot row vector for each row based on the categorical probability distribution defined by $\mathsf{softmax}(\mathbf{w})$. To evaluate this sample, we simply evaluate a forward pass through the network, treating the sampled sparse matrices from the softmax layers as the weights of linear layers:
\begin{equation}
\widetilde{\mathbf{h}}^{(l)} = \begin{bmatrix}\widetilde h_1^{(l)} \\ \vdots \\ \widetilde h_{M^{(l)}}^{(l)}\end{bmatrix} \equiv \mathsf{SAMPLE}\left(\begin{bmatrix} {\mathsf{softmax}(\mathbf{w}_1^{(l)})} \\ \vdots \\ {\mathsf{softmax}(\mathbf{w}_{M^{(l)}}^{(l)})}\end{bmatrix}\right)\mathbf{h}^{(l-1)},
\end{equation}

To complete the picture of the forward pass, we formalize how we deal with activations accepting multiple inputs. We define the action of the activation functions as follows:
\begin{equation}
\label{eq:argument-to-image}
h^{(l)}_i = \phi_i^{(l)}\left(\widetilde h_{j}^{(l)}, \dots, \widetilde h_{j + \alpha[\phi_i^{(l)}] - 1}^{(l)}\right),\;\;
j = \sum_{0 \leq k < i} \alpha\left[\phi_k^{(l)}\right].
\end{equation}

\subsection{OccamNet's Probability Distribution}
\label{sec:probability}
OccamNet parametrizes a probability distribution over all functions which it can sample. In particular, when OccamNet samples a function, it is really sampling a directed acyclic graph (DAG) which defines a computational path to compute a function. The probability of sampling a computational graph is equal to the product of the probabilities of the connections in the DAG which are connected to the output node.

Note that multiple computational graphs can correspond to the same function. In this paper, when we refer to a function sampled from OccamNet or the probability of a function according to OccamNet, we use function as a shorthand for a \textit{particular} computational graph corresponding to that function. Although this underspecifies the computational graph in question, this is never an issue because we always refer to functions in abstract.

When using OccamLlama for inference, we select the maximum probability function by sampling 100 functions from OccamNet, evaluating their probabilities as described above and selecting the maximum one.

\subsection{Initialization}
\label{sec:initialization}

This section describes how we calculate $\mathbf{W}^*$ from the main text. We wish to initialize $\mathbf{W}^*$ such that $p_{W^*}[f_1] = p_{W^*}[f_2]$ for all $f_1$ and $f_2$. Below, we assume that skip connections do not exist. However, the algorithm also works for skip connections, requiring only a small modification to Equation \ref{eq:argtoimgApp}.

Unfortunately, such an initialization is impossible for any OccamNet with two or more layers containing primitives with more than one argument. However, it is possible to initialize OccamNet such that a lower bound $q_{W^*}$ of the true probability $p_{W^*}$ is independent of $f.$

Define the probability of a function $f$ up to a given node as the product of the probabilities of the edges that lead to that node in the DAG of $f$. Intuitively, $q_W[f]$ approximates $p_W[f]$ by maintaining a lower bound on the probability of $f$ up to each node of an OccamNet and propagating that lower bound through the computational graph given by $f$.

To define $q_W$ more precisely, let $q_i^{(l)}[f]$ and $\widetilde q_i^{(l)}[f]$ be the probability bounds corresponding to the $i$th node of the $l$th image or arguments sublayer. We have suppressed the dependence on $W$ for notational convenience. We compute these probabilities starting with the inputs, for which we set $q_i^{(0)} = 1$. We then propagate probabilities to the arguments layers according to
\begin{equation}\label{eq:imtoargApp}
    \widetilde q_i^{(l+1)} = \mathsf{softmax}(\mathbf{w}_i^{(l+1)})_j q_j^{(l)},
\end{equation}
where $j$ is the node in the $l$th image layer which $f$ connects to the $i$th node of $(l+1)$th arguments layer. Similarly, we propagate probabilities to the image layers according to
\begin{equation}
\label{eq:argtoimgApp}
q_i^{(l)} = \prod_{k=n}^{n+\alpha[\phi_i^{(l)}]-1}\widetilde q_k^{(l)}, \hspace{10pt} n = \sum_{j=1}^{i-1} \alpha\left[\phi_j^{(l)}\right].
\end{equation}
Finally, we define $q_W[f] = q_0^{(L+1)}[f].$

In practice $q_W[f] \le p_W[f]$, where equality holds for many functions. In fact, $q_W[f] < p_W[f]$ only when part of the DAG of $f$ is used as input to two different arguments nodes. In cases such as these, the portion of the DAG that is used twice multiplicatively contributes the probability of its edges to $q_W[f]$ twice, artificially suppressing its value. However, because $q_W[f]$ is a lower bound, initializing $W^*$ to equalize $q_{W^*}$ still has the desired effect of ensuring adequate coverage for each $f$ in the initial probability distribution of OccamNet.

With this primer, we can now define the algorithm to initialize ${W^*}$ such that $q_{W^*}[f]$ is uniform.

The algorithm traverses through OccamNet layer by layer and establishes as an invariant that, after assigning the weights up to the $l$th layer, $\widetilde q_i^{(l)}[f]$ are equal for all $i$ and $f$. This implies that, after assigning the weights up to the $l$th layer, $q_i^{(l)}[f]$ are equal for all $f$, but not necessarily for all $i$. We denote the common value of $\widetilde q_i^{(l)}[f]$ as $\widetilde q^{(l)}$ and the common value of $q_i^{(l)}[f]$ as $q_i^{(l)}.$

The algorithm starts with input layer, where $q_i^{(0)} = 1$ automatically. Once the invariant above is true for a given $l,$ the algorithm sets
\begin{equation}\label{eq:weight_init}
    \left(\mathbf{w}_i^{*(l+1)}\right)_j = \log\left(\frac{\min_k\left(q_k^{(l)}\right)}{q_j^{(l)}}\right)
\end{equation} for all $i,j$, where $\left(\mathbf{w}_i^{*(l+1)}\right)_j$ denotes the weight connecting the $j$th node in the $l$th image layer to the $i$th node in the $(l+1)$th arguments layer. This establishes the invariant for $l+1$ because 
\begin{align*}
    q_j^{(l)} \mathsf{softmax}(\mathbf{w}_i^{*(l+1)})_j &= \frac{q_j^{(l)} \exp\left[\left(\mathbf{w}_i^{*(l+1)}\right)_j\right]}{\sum_k \exp\left[\left(\mathbf{w}_i^{*(l+1)}\right)_k\right]}\\
    &= \frac{q_j^{(l)} \min_k\left(q_k^{(l)}\right)/q_j^{(l)}}{\sum_k \min_m\left(q_m^{(l)}\right)/q_k^{(l)}}\\
    &= \frac{1}{\sum_k 1/q_k^{(l)}},
\end{align*}
which is a constant over both $i$ and $j$, so $\widetilde q_i^{(l+1)}[f]$ is a constant over both $i$ and $f.$ The algorithm repeats the above procedure until it has traversed the entire network.

In summary, the algorithm involves the following steps:
\begin{enumerate}
    \item Set $l=0$ and $q_i^{(l)}=1$.
    \item Increment $l$ by 1.
    \item Set $\mathbf{W}^{*(l)}$ according to Equation \ref{eq:weight_init}.
    \item If $l < L+1$, Compute $\widetilde q^{(l+1)}$ and $q_i^{(l+1)}$.
    \item Return to step 2 until $l = L+1.$
\end{enumerate}